\documentclass[conference]{IEEEtran}
\usepackage{cite}
\usepackage{amsmath,amssymb,amsfonts}
\usepackage{graphicx}
\usepackage{array}
\usepackage{booktabs}
\usepackage{float}
\usepackage{subcaption}
\usepackage{textcomp}
\usepackage{url}
\usepackage{xcolor}
\usepackage{listings}
\lstdefinestyle{prompt}{
  basicstyle=\small\ttfamily,
  backgroundcolor=\color{gray!10},
  frame=single,
  rulecolor=\color{gray!40},
  breaklines=true,
  breakatwhitespace=true,
  columns=fullflexible,
  aboveskip=6pt,
  belowskip=6pt
}
\usepackage{soul}
\usepackage{comment}
\usepackage[colorinlistoftodos,textsize=small]{todonotes}
\usepackage{orcidlink}

\begin{document}
    \title{NAVI-Orbital: First In-Orbit Demonstration of a Zero-Shot Vision-Language Model for Autonomous Earth Observation}

    \author{
        \IEEEauthorblockN{
            Juan M. Delfa Victoria\IEEEauthorrefmark{1},\orcidlink{0009-0004-9689-9424},
            Taran Cyriac John\IEEEauthorrefmark{1}\IEEEauthorrefmark{2},\orcidlink{0000-0003-4230-4857}
            Andrew W. Herson\IEEEauthorrefmark{3}
        }
        \IEEEauthorblockA{
            \IEEEauthorrefmark{1}\textit{Jet Propulsion Laboratory}\\
            \textit{California Institute of Technology}\\
            Pasadena, CA, USA \\
            Email: juan.m.delfa.victoria@jpl.nasa.gov, taran.john@jpl.nasa.gov
        }
        \IEEEauthorblockA{
            \IEEEauthorrefmark{2}\textit{School of Engineering and Computer Science}\\
            \textit{Victoria University of Wellington}\\
            Wellington 6140, New Zealand\\
            Email: taran.john@vuw.ac.nz
        }
        \IEEEauthorblockA{
            \IEEEauthorrefmark{3}\textit{Loft Orbital}\\
            \textit{321 11th St, San Francisco, CA 94103}\\
            Email: andrew.herson@loftorbital.com
        }
    }

    \maketitle
    \pagestyle{plain}
    \pagenumbering{arabic}

    \begin{abstract}
    As Earth Observation data generation outpaces downlink bandwidth and human-in-the-loop processing, a widening gap has emerged between onboard collection and actionable ground intelligence.     
    This paper presents NAVI-Orbital,
    a software system deployed on a Low Earth Orbit (LEO)
    spacecraft. On April 16, 2026, NAVI-Orbital achieved what is, to the authors' knowledge, the first in-orbit demonstration of a vision-language model performing autonomous multi-modal inference entirely onboard. NAVI-Orbital uses a local vision-language model
    (Gemma~3) to classify each captured scene, produce a text
    description of its content and the relationships between its features,
    and respond to operator follow-up via natural-language dialogue. The
    system is re-tasked through plain-English prompts in place of
    conventional command sequences, and is orchestrated by a graph-based
    state machine (LangGraph) coordinating dedicated agents for detection
    and dialogue. Results across ground benchmarking (88.16\% accuracy on
    the 7{,}960-image curated AID benchmark), Flatsat validation, and live in-orbit captures
    of newly acquired, previously unseen Earth imagery (including uncorrected YAM-9 imagery, processed onboard with
    hardware-accelerated GPU inference and no fine-tuning for the flight
    instrument) demonstrate the feasibility of running foundation models
    on satellite-class edge computers to invert the conventional
    acquire-then-downlink-everything bandwidth profile through semantic
    compression of Earth observations in-orbit.
    \end{abstract}


    \section{Introduction}
    \label{sec:intro}
    Modern Earth Observation (EO) instruments are generating data at rapidly increasing rates, yet the two critical pathways for managing that volume, physical downlink bandwidth and human-in-the-loop review capacity, cannot scale proportionally. This creates a widening gap between onboard data generation and exploited information available to ground operators \cite{gomez2024tackling}.
    
    Onboard processing offers a partial remedy, but current implementations are largely limited to ``specialist
    detectors'': algorithms trained to identify specific, pre-defined pixel patterns (e.g., drawing a bounding box
    around a ship) \cite{giuffrida2021phisat}. These systems are effective within their narrow scope, but adapting
    them to new phenomenologies requires retraining models, validating new binaries, and executing complex software
    updates, a process that is slow, expensive, and fundamentally at odds with the agility that future missions
    demand.

    \subsection{NAVI-Orbital}
    NAVI-Orbital is a software framework developed by researchers at NASA Jet Propulsion Laboratory (JPL) and deployed onboard Loft Orbital’s YAM-9 Low Earth Orbit (LEO) spacecraft that uses a
    multi-modal large language model (Google's Gemma~3 \cite{google2025gemma}) to understand the content of images and
    describe it in plain English. By jointly processing visual and textual information, the model develops a
    semantic understanding of scene content, producing contextualized descriptions that identify not just
    \textit{what} objects and features are present in an image but also the \textit{relationships} between them
    (e.g., ``a highway cutting through a forested area''). Unlike traditional specialist detectors that require
    retraining for each new target class, NAVI-Orbital leverages the open-vocabulary nature of
    vision-language models~\cite{Zareianetal2021, Radfordetal2021}, enabling it to adapt to new observation tasks
    through prompt modifications alone, without altering the underlying model architecture
    (Section~\ref{NAVIArchitecture}). This capability will enable future satellites to react
    intelligently to their environment rather than operating as passive sensors responding to predetermined
    triggers.

    The system operates through a sequence of steps, from image reception to human-spacecraft dialogue. A user
    uploads a plain-English prompt defining what to look for and a list of target labels. NAVI classifies each
    image, generates a text description of the scene, and stores a structured record for every image. Compact
    text summaries are downlinked to the ground, where operators review them and request full images only for
    high-value scenes. Provided the spacecraft has direct communication with Earth, a human operator can further
    interrogate the results through scripted question sets or interactive chat.

    This pipeline is orchestrated by a multi-agent architecture composed of three self-contained agents that hand
    off work to each other: an \textit{orchestrator} that coordinates execution, a \textit{detector} that
    analyzes, classifies, and summarizes images, and a \textit{dialogue} agent that enables operators to ask
    questions about the results. This design makes NAVI adaptable to different missions without rebuilding from
    scratch.

    \subsection{Contributions}
    The primary contributions are:

    \begin{itemize}
        \item \textbf{First In-Orbit Multi-Modal Inference:}
        NAVI actively processes fresh, live imagery directly from the spacecraft's sensors. The Gemma~3 model runs with hardware-accelerated GPU inference on a satellite-class edge processor under strict size, weight, and power constraints, demonstrating that complex, multi-agent AI workflows can execute on edge computing hardware onboard a spacecraft.

        \item \textbf{Generalist Knowledge Without Retraining:}
        Large-scale models trained on internet-scale datasets possess generalized recognition capabilities that
        eliminate the need for individualized retraining for every specific concept or instrument. NAVI leverages
        this zero-shot capability to adapt to new observation tasks by changing text prompts alone, without
        altering the core software or deploying new neural network architectures. The system was validated across
        approximately 7{,}000 images spanning multiple datasets and multiple hardware platforms, and successfully
        processed live YAM-9 imagery without any specialized fine-tuning for that optical instrument, including uncorrected live imagery.

        \item \textbf{Multi-Modal Contextual Reasoning:}
        By leveraging a Vision-Language Model (VLM), NAVI develops a semantic understanding of observed scenes. Traditional onboard classifiers output only discrete
        labels or bounding boxes with no contextual awareness. NAVI instead produces rich, contextualized
        descriptions, enabling knowledge-driven autonomy where the spacecraft interprets its environment rather
        than merely detecting predetermined targets.

        \item \textbf{Plain-English Operations:}
        NAVI replaces complex spacecraft commanding and data retrieval with natural language for the entire
        input/output loop. On the uplink, operators define what to look for using simple text prompts, specifying
        target labels and processing instructions rather than rigid command sequences. On the downlink, the spacecraft returns plain-English text summaries of every image it
        processes. Operators can further interrogate results through scripted question sets or interactive chat,
        selectively downlinking only high-value imagery and thereby optimizing limited communication
        windows \cite{gomez2024tackling}. When an Intersatellite Link (ISL) is available, this dialogue can occur
        in near real-time, transforming the entire interaction loop with the satellite into an intuitive, natural
        conversation.
    \end{itemize}

    \subsection{Paper Organization}
    The remainder of this paper is organized as follows. Section~II surveys related work in onboard AI, generative
    models on manned platforms, and vision-language models in remote sensing. Section~III describes the system
    architecture, including the LangGraph-based conductor graph, agent ecosystem, and hardware integration.
    Section~IV details the experimental setup, datasets, and phased validation approach. Section~V presents results
    from ground benchmarking, Flatsat validation, and in-orbit demonstration. Sections~VI and~VII discuss
    implications, limitations, and conclusions.

    \section{Related Work}

    \subsection{Onboard AI for Earth Observation}
    The transition from ground-in-the-loop operations to onboard autonomy has been driven by the need to maximize spacecraft science utility. The Autonomous Sciencecraft Experiment (ASE) on EO-1 pioneered the use of onboard classifiers to detect features like volcanic activity and trigger
    subsequent data acquisitions \cite{chien2005eo1}. As space-rated computational hardware matured, the focus shifted toward Deep Neural Networks. The $\Phi$-Sat-1 mission demonstrated the efficacy of dedicated hardware accelerators for filtering cloud-covered imagery
    using Convolutional Neural Networks (CNNs) \cite{giuffrida2020cloudscout}.

    While highly effective, these systems rely on fixed-class supervision. Adapting them to new phenomenologies traditionally required complex, monolithic software updates. To address this limitation, recent missions have validated re-trainable AI architectures in orbit. Early work demonstrated the feasibility of machine learning for global flood mapping on resource-constrained satellite hardware \cite{mateogarcia2021towards}, establishing the foundation for adaptive onboard segmentation. This approach was later extended to orbital deployment with the in-orbit demonstration of a re-trainable machine learning payload aboard D-Orbit's ION satellite, where segmentation models were continuously updated post-launch to maintain high accuracy for dynamic flood detection tasks \cite{MateoGarciaetal2023}. Concurrently, the European Space Agency's OPS-SAT nanosatellite has served as a dedicated orbital testbed, allowing diverse deep learning algorithms (such as the SmartCam application) to be dynamically uploaded, hybridized, and executed on edge hardware in space \cite{LabrecheMladenov2023}. Pushing the boundaries of data reduction even further, missions like KP Labs' Intuition-1 \cite{Wijata2024} and HYPSO-1 \cite{justo2025hyperspectral} have recently utilized onboard DNNs to process hyperspectral data directly in orbit, converting raw instrument data into actionable intelligence and substantially decreasing the volume of telemetry transmitted to the ground. Scaling beyond single algorithms, contemporary architectures now support concurrent, multi-application processing; Wuhan University's Luojia3 leverages heterogeneous parallel computing for simultaneous application execution \cite{zhang2022expandable}, and ESA's $\Phi$-Sat-2 executes complete end-to-end data processing chains natively in orbit.
    
    Beyond static image processing, Dynamic Targeting (DT) extends onboard autonomy by utilizing a ``lookahead'' sensor to analyze the scene ahead of the satellite's ground track. This enables the primary instrument to autonomously slew and capture high-value targets with elevated precision. Recent flight demonstrations on the CogniSAT-6 spacecraft have validated DT in Low Earth Orbit (LEO) environments, successfully utilizing edge computing to perform lookahead analysis and drive primary sensor tasking in real-time \cite{dt-spaceops-2025}.
    
    However, despite these significant advancements, current implementations generally rely on specialized, pre-trained classifiers customized for specific targets (e.g., clouds, floods, or thermal anomalies) rather than open-vocabulary interpretation.

    \subsection{Generative AI in Space}
    Parallel to advances in autonomous remote sensing, early efforts have emerged in deploying Large Language Models (LLMs) to support both manned spaceflight and autonomous subsystem control. In 2024, a text-based generative AI model was deployed aboard the International Space Station (ISS) utilizing the HPE Spaceborne Computer-2 infrastructure. This system was designed to assist astronauts with maintenance procedures through Retrieval-Augmented Generation (RAG) \cite{boozallen2024generative}. 
    
    In 2025, this paradigm advanced with ``Space Llama,'' a collaboration between Meta and Booz Allen that successfully deployed a quantized Llama 3.2 model to the ISS \cite{meta2025spacellama}. Functioning as a multimodal digital assistant, Space Llama facilitates scientific and technical tasks, including predictive maintenance and autonomous access to documentation, effectively replacing physical space manuals for the crew. 
    
    While Space Llama exemplifies the utility of LLMs as interactive aides, recent research has transitioned toward active agentic supervision. The ASTREA (Agentic System for Thermal Regulation and Embedded Adaptation) mission represents the first agentic system executed on flight-heritage hardware for autonomous operations, with in-orbit validation aboard the ISS \cite{mousist2025astrea}. ASTREA utilizes an asynchronous hybrid architecture wherein a resource-constrained LLM provides strategic, semantic guidance to a real-time reinforcement learning controller tasked with orbital thermal regulation \cite{mousist2025astrea}. Ground and in-orbit experiments demonstrated that this semantic reasoning capability improves thermal stability, extends episode durations, and optimizes CPU utilization, demonstrating the viability of combining LLMs with adaptive control under strict hardware constraints.

    \subsection{Key Technologies for Generative AI in Space}

    \subsubsection{Edge Inference and Hardware Optimization}
    The deployment of multi-billion parameter foundation models onto constrained spacecraft buses necessitates extensive hardware abstraction and memory optimization. The \texttt{llama.cpp} framework provides a dependency-free, C/C++ inference engine that interacts directly with edge CPUs and neural processing units (NPUs) \cite{gerganov2023llamacpp}. To mitigate severe memory bandwidth bottlenecks, researchers increasingly employ block-wise quantization techniques to reduce tensor precision \cite{dettmers2022optimizers}. Recent unified evaluations of quantization formats confirm that sub-byte integer representations significantly reduce memory footprints while preserving complex reasoning and instruction-following capabilities \cite{kurt2026quantization}.
    
    \subsubsection{Vision-Language and Geospatial Foundation Models}
    Generative models deployed in orbit must be tailored to comprehend the unique spatial and spectral characteristics of Earth observation telemetry. NAVI-Orbital uses the Gemma~3~4B instruction-tuned model in its Q4\_0 GGUF quantization with a dedicated multimodal projector (\texttt{mmproj}) for vision--language
    fusion. This model introduces hybrid interleaved attention, significantly reducing the Key-Value (KV) cache memory overhead required to process long contexts of up to 128K tokens \cite{google2025gemma}. It also utilizes a dynamic ``Pan and Scan'' visual integration strategy to preserve critical high-resolution metadata without destructive down-sampling \cite{google2025gemma}. Within the remote sensing domain specifically, RemoteCLIP pioneered contrastive alignment to bridge textual descriptions and orbital imagery, enabling open-vocabulary, zero-shot classification and retrieval \cite{liu2024remoteclip}. GeoChat expanded these capabilities by introducing grounded spatial reasoning, allowing models to process and output precise bounding-box coordinates interleaved with natural language \cite{kuckreja2024geochat}. Addressing the large variance in satellite imagery, RSCoVLM implements a unified dynamic-resolution strategy and multi-task learning (MTL), evaluating generative outputs against strict object detection metrics \cite{li2026rscovlm}. At the most comprehensive scale, TerraMind demonstrates the efficacy of large-scale modality integration, processing both pixel-level and token-level representations across diverse geospatial data types \cite{jakubik2025terramind}.
    
    \subsubsection{Agentic Orchestration Frameworks}
    Translating probabilistic outputs from vision-language models into deterministic spacecraft commands requires robust supervision. Recent surveys on multi-agent orchestration highlight the necessity of structured communication protocols and state management to ensure safety in autonomous systems \cite{zhu2026orchestration}. Taxonomies of agentic AI distinguish between simple linear chains and complex, stateful graph orchestrations, emphasizing the latter's superiority for applications requiring human-in-the-loop oversight and automated error recovery \cite{sapkota2025langchain}. Frameworks such as LangGraph provide this essential low-level orchestration by modeling multi-agent interactions as directed cyclic graphs with persistent memory \cite{wang2024agent}. This structure ensures that cognitive tasks are safely compartmentalized, repeatedly validated, and preserved across hardware reboots before interacting with critical flight software.
    
    \subsection{Positioning NAVI-Orbital}
    The operational paradigm of NAVI-Orbital differs from these prior generative AI deployments in three fundamental ways. First, unlike systems hosted within the pressurized environment of the ISS \cite{boozallen2024generative, meta2025spacellama, mousist2025astrea}, NAVI-Orbital is deployed directly onboard an autonomous Earth observation spacecraft. Second, rather than relying on rack-mounted server infrastructure, it executes entirely on a power-constrained ARM-based edge computing platform. Third, while early spaceborne LLMs focus primarily on text-based assistance or internal subsystem regulation, NAVI-Orbital focuses on multi-modal image semantic understanding. By natively processing visual data, it bridges the gap between external perception and semantic reasoning, autonomously translating raw remote sensing imagery into contextualized information without human intervention.

    \begin{figure*}[!t]
        \centering
        \includegraphics[width=\textwidth]{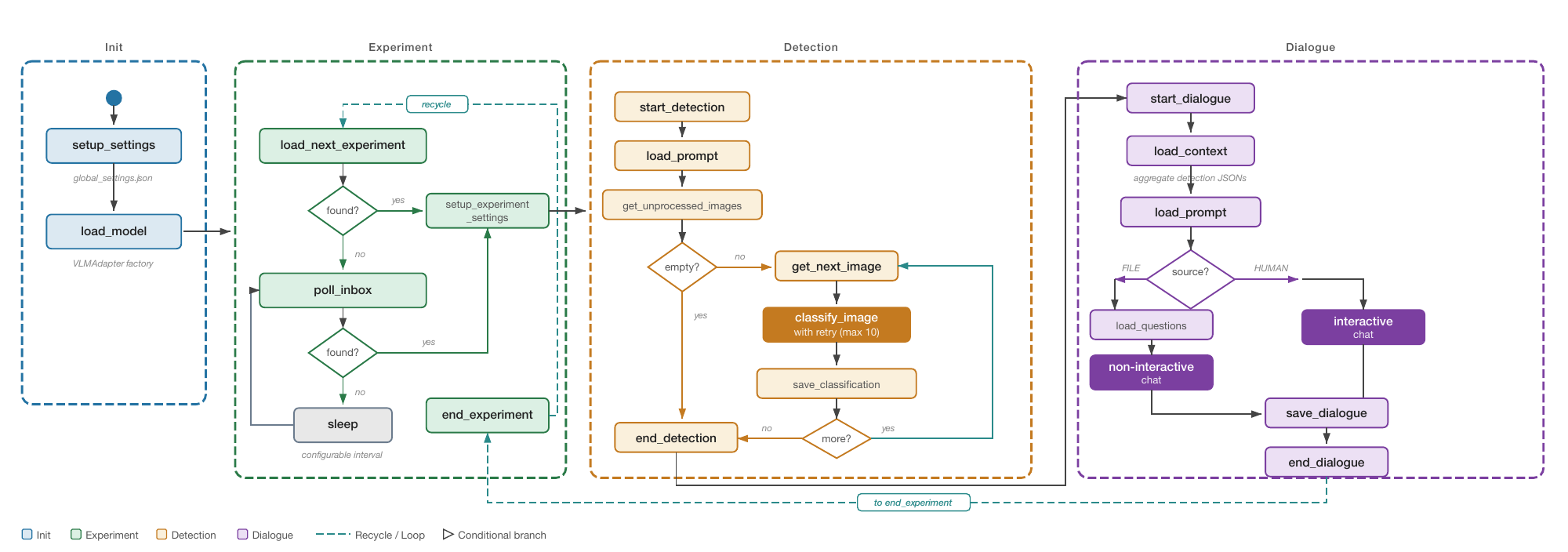}
        \caption
        {The Conductor Graph: a directed state graph composed of four sub-graphs. Rounded rectangles represent LangGraph
        nodes; dashed borders group nodes by sub-graph. Diamond shapes denote conditional edges. The outer loop recycles
        from \texttt{end\_experiment} back to \texttt{load\_next\_experiment} for continuous autonomous operation.}
        \label{fig:conductor_graph}
    \end{figure*}

    \section{System Architecture}

    \subsection{The YAM-9 Platform}
     NAVI-Orbital runs on Loft Orbital’s YAM-9 satellite, a multi-mission platform flying a heterogeneous compute cluster composed of multiple edge-class processing elements designed for operation in harsh LEO environments. Its open compute architecture allows the deployment and operation of custom software and foundation models, and builds on Loft’s existing in-orbit compute infrastructure.  
    
    NAVI-Orbital was originally optimized for a specific ARM/GPU execution environment. The system architecture targets high-performance edge processors commonly used in modern smart satellites.  Platform variants include Qualcomm's Snapdragon, NVIDIA's Orin-series devices, and similar devices capable of hosting Gemma~3 4B (requires a minimum of 8\,GB of VRAM).  Since its initial deployment, however, the application has been generalized to operate across multiple compute elements, enabling flexible task distribution and broader portability.

    \subsection{NAVI-Orbital Software Architecture}
    \label{NAVIArchitecture}
    NAVI-Orbital implements a hierarchical, agentic architecture designed to operate autonomously within the
    constraints of edge-computing hardware on a spacecraft.

    The core logic is orchestrated by a state machine implemented using \texttt{LangGraph}~
    \cite{sapkota2025langchain, langchain2024langgraph}, which manages the flow of control between three distinct
    functional layers: the Conductor (Orchestration), the Detector (Reasoning), and the Dialogue (Interaction) agents.
    During architecture evaluation, the Anthropic Model Context Protocol (MCP) was
    considered as an alternative orchestration framework. However, LangGraph was ultimately selected because its graph-based state machine provides deterministic, verifiable transitions between operational phases. This was a key requirement for onboard applications where predictable behavior and state traceability are essential for autonomous spacecraft operations.
    The underlying vision-language model and NAVI-Orbital's prompt-based architecture are inherently open-vocabulary~
    \cite{Zareianetal2021, Radfordetal2021}, capable of recognizing and describing arbitrary
    visual concepts without task-specific retraining. However, for operational deployment, each experiment is configured
    with a constrained label set passed in the detection prompt (Section~\ref{sec:workflow}
    ). This design choice trades some of the model's generality for output determinism: by restricting the model to a
    predefined vocabulary, the system ensures that
    classification results are parseable by downstream spacecraft subsystems and compatible with the regex-based
    validation gate in the Detector Agent's retry loop (Section~\ref{NAVIArchitecture}). This
    hybrid approach preserves the model's zero-shot adaptability (new label sets require only a prompt edit, not
    model retraining) while maintaining the deterministic output semantics required for
    autonomous operation.
    
    \subsubsection{Core Layer}
    The Core Layer provides the foundation that every agent builds on: a control plane, a bus-facing I/O interface, model abstractions, and shared utilities.

    \paragraph{Control Plane}
    A FastAPI service (\texttt{api.py}) exposes lifecycle endpoints (\texttt{/start}, \texttt{/stop}, \texttt{/kill}), configuration endpoints (\texttt{/get-config}, \texttt{/reload-config}), and a \texttt{/health} probe reporting workflow status, accelerator availability, and path validity. The agent graph is spawned as a separate \texttt{multiprocessing.Process}, so the control plane remains responsive and can reclaim GPU memory by terminating the worker without restarting the service.

    \paragraph{Inbox/Outbox Interface}
    All interaction with the spacecraft bus is mediated by two filesystem queues. An external producer deposits a session descriptor (a JSON file listing a \texttt{session\_id} and target images) into \texttt{io/inbox/}; the Conductor polls this directory once preloaded and template-based experiments are exhausted (Section~\ref{NAVIArchitecture}). On completion, the Conductor writes an aggregated result JSON containing detections, dialogue transcript, and log excerpts to \texttt{io/outbox/}, where it is retrieved by the bus. This file-based contract decouples NAVI-Orbital from any specific spacecraft communication protocol and matches the asynchronous, store-and-forward nature of typical satellite telecommand and telemetry links.

    \paragraph{Model Abstraction}
    A \texttt{VLMAdapter} abstract base class is implemented by \texttt{GemmaVLAdapter} (HuggingFace pipeline) and \texttt{LlamaCppAdapter} (GGUF via \texttt{llama.cpp} with a multimodal projector). A factory selects the concrete adapter from \texttt{model\_settings.json}, and each adapter negotiates a device-appropriate precision (4-bit, 8-bit, \texttt{bfloat16}, \texttt{float16}, or \texttt{float32}) with a fallback chain tuned per CUDA, MPS, and CPU. For HuggingFace models, runtime quantization is applied via \texttt{bitsandbytes} (4-bit or 8-bit); for GGUF models, quantization is embedded into the model file at export time (e.g., Q4\_0), and the adapter loads the pre-quantized weights directly.

    \paragraph{Utilities and Configuration}
    A shared \texttt{file\_handler} module owns experiment discovery, image scanning, and result serialization; a logging module provides dual file/console output with per-experiment archival; and a context-manager utility redirects C-level \texttt{stdout}/\texttt{stderr} from \texttt{llama.cpp} into the experiment log so native engine traces are preserved alongside Python-level events. Two JSON files anchor configuration: \texttt{global\_settings.json} (paths only) and \texttt{model\_settings.json} (adapter selection, weights, device, precision, and inference hyperparameters), the latter hot-reloadable via the API while the workflow is idle.

    \subsubsection{Agents Layer}
    The architectural backbone of NAVI-Orbital is the \textbf{Conductor Graph}, a directed state graph implemented in
    \texttt{LangGraph}
    that enforces deterministic transitions between operational phases. The graph is assembled from four composable
    sub-graphs, each encapsulating a distinct phase of the processing pipeline (Fig.~\ref{fig:conductor_graph}):

    \begin{enumerate}
        \item \textbf{Init Sub-graph:} Loads global configuration (\texttt{global\_settings.json}
        ), instantiates the VLM adapter, and transitions to the experiment phase.
        \item \textbf{Experiment Sub-graph:}
        Implements a three-stage experiment loading priority: (i)~existing unprocessed experiment folders, (ii)~mandatory
        template-based experiments (e.g., preloaded datasets), and (iii)~live inbox polling with a configurable sleep
        interval. Each experiment run is assigned a unique UUID. This sub-graph contains conditional edges that route to either experiment setup or a poll-sleep loop
        when no work is available.
        \item \textbf{Detection Sub-graph:}
        Manages the image processing pipeline: loading prompts, iterating through the image queue, invoking the
        classify-with-retry loop, and serializing results. A conditional edge after each classification checks whether
        images remain in the queue, looping back or terminating the phase accordingly.
        \item \textbf{Dialogue Sub-graph:}
        Aggregates detection results into a context window, then branches conditionally into either interactive (live
        chat) or non-interactive (file-based batch) dialogue mode based on the experiment configuration.
    \end{enumerate}

    The system tracks its operational phase via a \texttt{Status} enumeration (\texttt{IDLE}, \texttt{TRANSITION},
    \texttt{DETECTION}, \texttt{DIALOGUE}
    ), which is orthogonal to the graph nodes and is used for logging and state persistence. Upon completing all phases,
    the graph cycles back from \texttt{end\_experiment} to \texttt{load\_next\_experiment}
    , creating a continuous autonomous processing loop.

    \section{Experimental Setup}

    \subsection{Workflow}
    \label{sec:workflow}
    NAVI-Orbital operates as a persistent service. Once started, the engine enters an idle state and awaits a
    JSON experiment descriptor that specifies the image source, label set, detection prompt, and dialogue
    configuration. Upon receipt, the Conductor Graph (Section~\ref{NAVIArchitecture}) drives execution through the Detection and Dialogue sub-graphs described in Section~\ref{NAVIArchitecture}, then cycles back to the experiment loader. The classify-with-retry loop is configured for up to 3 attempts during ground evaluation and up to 10 attempts during flight.

    Two experiment modes govern how images are sourced:
    \begin{itemize}
        \item \textbf{Preloaded Mode (Validation):}
        The system ingests static datasets (e.g., Google AID, Sentinel-2) from a local directory to benchmark model
        performance against known ground truth.
        \item \textbf{Live Mode (Flight):}
        The system polls the inbox for new imagery captured by the spacecraft bus, processing files
        in real-time as they are written to disk.
    \end{itemize}

    \subsection{Datasets}
    \label{sec:datasets}
    \subsubsection{Google AID}
    \label{sec:datasetgoogle}
    The primary evaluation dataset is derived from the Aerial Image Dataset (AID) \cite{xia2017aid}, a large-scale
    benchmark comprising approximately 10{,}000 images across 30 aerial scene classes at 600$\times$600~px resolution.
    We curated AID to 18 classes totaling 7{,}960 images through two complementary strategies.
    First, \textit{semantic deduplication} collapsed classes with overlapping visual semantics: Dense Residential,
    Medium Residential, and Sparse Residential were merged into a single Residential class; Meadow was absorbed into
    Agricultural; BareLand was subsumed by Desert; Park by Forest; and Viaduct by Bridge. Stadium was relabeled to
    Venue to better reflect operational label semantics.
    Second, a \textit{mission-relevance filter} removed seven classes whose defining features are sub-pixel-scale
    and therefore unlikely to be discriminable from LEO optical instruments: BaseballDiamond, Church, School, Center,
    Resort, Square, and Playground. The resulting 18-class subset retains the diversity of the original benchmark while
    reflecting labels that a satellite-borne classifier would plausibly encounter.

\begin{figure*}[!htb]
        \centering
        %% -------- Group A: Google AID (9×2) --------
        \textbf{\scriptsize (a) Google AID~\cite{xia2017aid} --- 18 classes}\\[2pt]
        %% Row 1
        \begin{subfigure}[b]{0.105\textwidth}\centering\includegraphics[width=\linewidth]{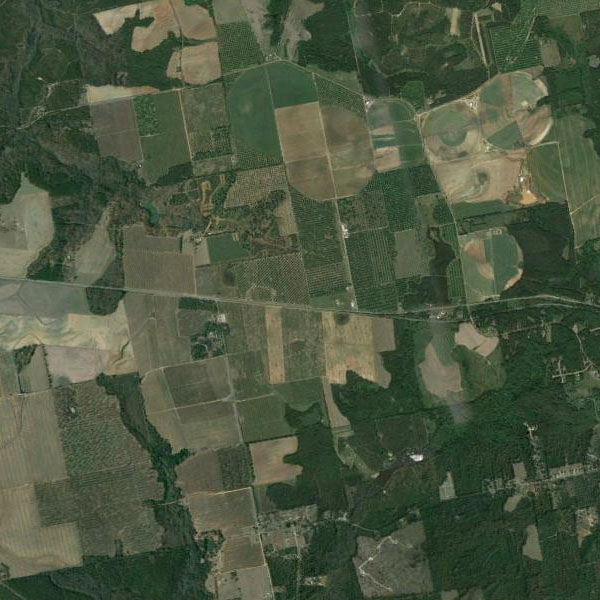}\\[-2pt]{\tiny Agricultural}\phantomcaption\label{fig:aid-agricultural}\end{subfigure}\hfill
        \begin{subfigure}[b]{0.105\textwidth}\centering\includegraphics[width=\linewidth]{images/preloaded-aid/ap_2.jpg}\\[-2pt]{\tiny Airport}\phantomcaption\label{fig:aid-airpot}\end{subfigure}\hfill
        \begin{subfigure}[b]{0.105\textwidth}\centering\includegraphics[width=\linewidth]{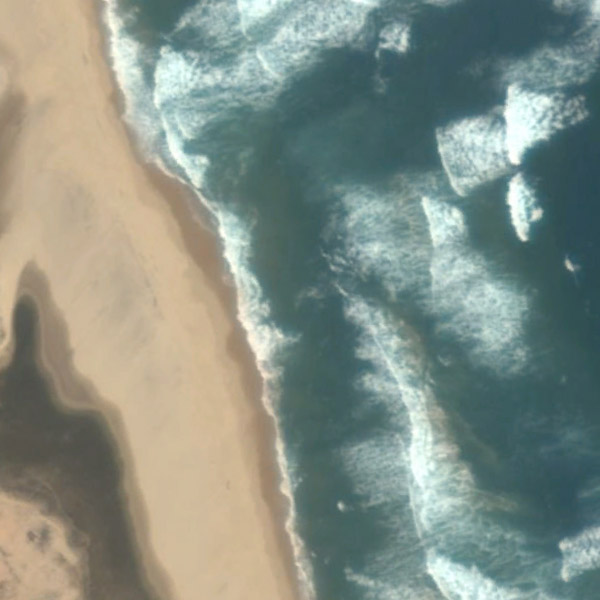}\\[-2pt]{\tiny Beach}\end{subfigure}\hfill
        \begin{subfigure}[b]{0.105\textwidth}\centering\includegraphics[width=\linewidth]{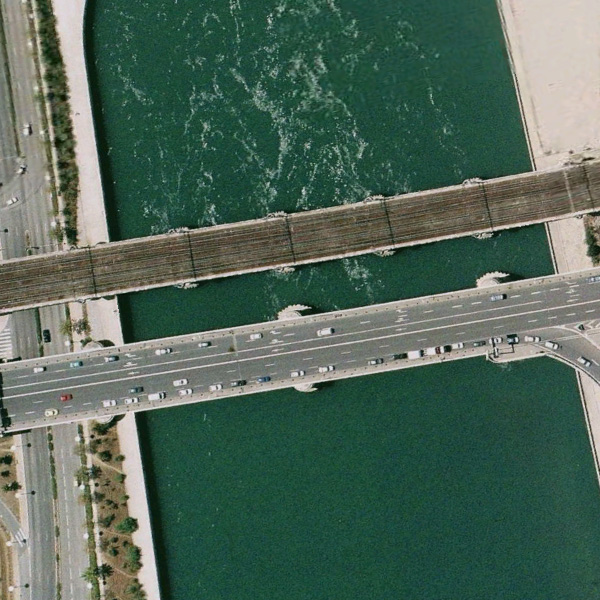}\\[-2pt]{\tiny Bridge}\end{subfigure}\hfill
        \begin{subfigure}[b]{0.105\textwidth}\centering\includegraphics[width=\linewidth]{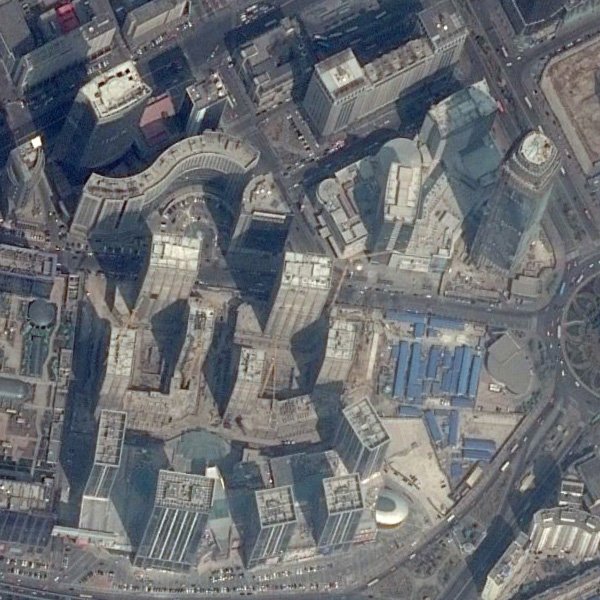}\\[-2pt]{\tiny Comm.\ Dist.}\end{subfigure}\hfill
        \begin{subfigure}[b]{0.105\textwidth}\centering\includegraphics[width=\linewidth]{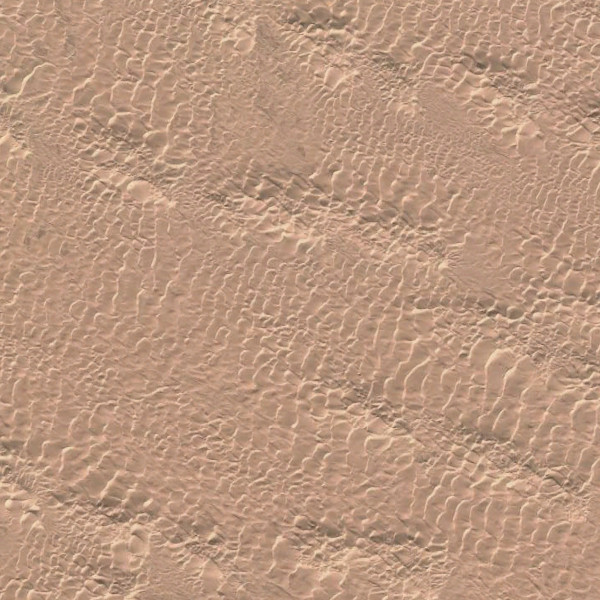}\\[-2pt]{\tiny Desert}\phantomcaption\label{fig:aid-desert}\end{subfigure}\hfill
        \begin{subfigure}[b]{0.105\textwidth}\centering\includegraphics[width=\linewidth]{images/preloaded-aid/fr_6.jpg}\\[-2pt]{\tiny Forest}\phantomcaption\label{fig:aid-forest}\end{subfigure}\hfill
        \begin{subfigure}[b]{0.105\textwidth}\centering\includegraphics[width=\linewidth]{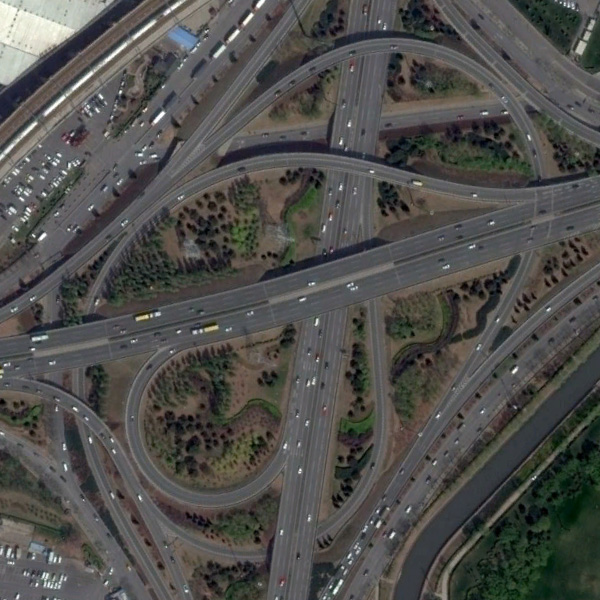}\\[-2pt]{\tiny Hwy.\ Intchg.}\end{subfigure}\hfill
        \begin{subfigure}[b]{0.105\textwidth}\centering\includegraphics[width=\linewidth]{images/preloaded-aid/id_1.jpg}\\[-2pt]{\tiny Industrial}\end{subfigure}
        \\[2pt]
        %% Row 2
        \begin{subfigure}[b]{0.105\textwidth}\centering\includegraphics[width=\linewidth]{images/preloaded-aid/mt_3.jpg}\\[-2pt]{\tiny Mountain}\end{subfigure}\hfill
        \begin{subfigure}[b]{0.105\textwidth}\centering\includegraphics[width=\linewidth]{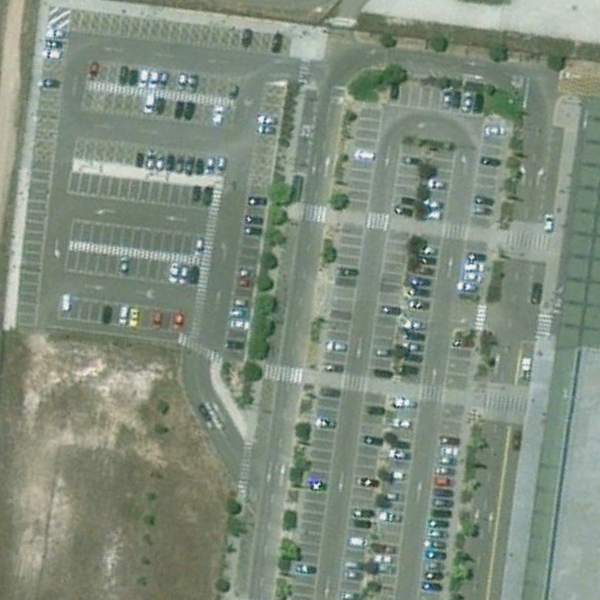}\\[-2pt]{\tiny Parking}\end{subfigure}\hfill
        \begin{subfigure}[b]{0.105\textwidth}\centering\includegraphics[width=\linewidth]{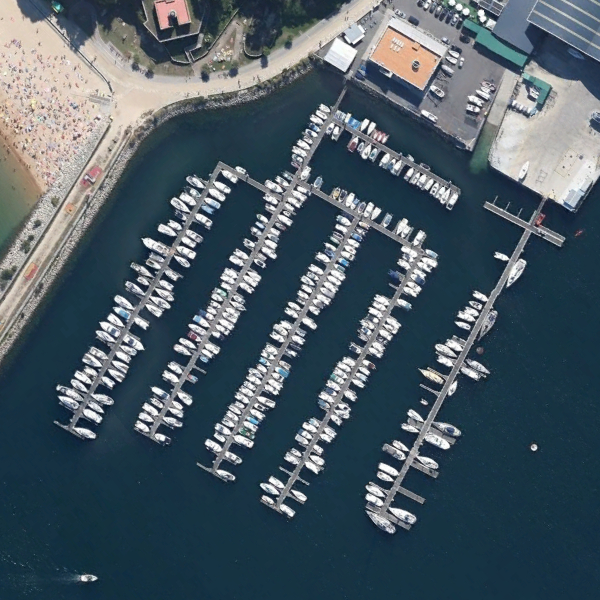}\\[-2pt]{\tiny Port}\phantomcaption\label{fig:aid-port}\end{subfigure}\hfill
        \begin{subfigure}[b]{0.105\textwidth}\centering\includegraphics[width=\linewidth]{images/preloaded-aid/res.jpg}\\[-2pt]{\tiny Residential}\end{subfigure}\hfill
        \begin{subfigure}[b]{0.105\textwidth}\centering\includegraphics[width=\linewidth]{images/preloaded-aid/rv_6.jpg}\\[-2pt]{\tiny River}\end{subfigure}\hfill
        \begin{subfigure}[b]{0.105\textwidth}\centering\includegraphics[width=\linewidth]{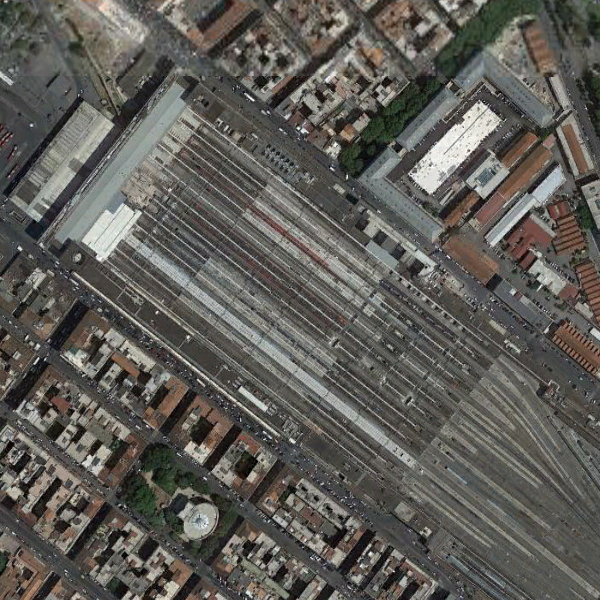}\\[-2pt]{\tiny Railway Stn.}\end{subfigure}\hfill
        \begin{subfigure}[b]{0.105\textwidth}\centering\includegraphics[width=\linewidth]{images/preloaded-aid/st_2.jpg}\\[-2pt]{\tiny Storage Tanks}\end{subfigure}\hfill
        \begin{subfigure}[b]{0.105\textwidth}\centering\includegraphics[width=\linewidth]{images/preloaded-aid/vn_31.jpg}\\[-2pt]{\tiny Venue}\end{subfigure}\hfill
        \begin{subfigure}[b]{0.105\textwidth}\centering\includegraphics[width=\linewidth]{images/preloaded-aid/wr_59.jpg}\\[-2pt]{\tiny Water Res.}\end{subfigure}

        \vspace{4pt}
        %% -------- Group B: Sentinel-2 (7×1) --------
        \textbf{\scriptsize (b) Sentinel-2 --- ESA WorldCover taxonomy}\\[2pt]
        \begin{subfigure}[b]{0.13\textwidth}\centering\includegraphics[width=\linewidth]{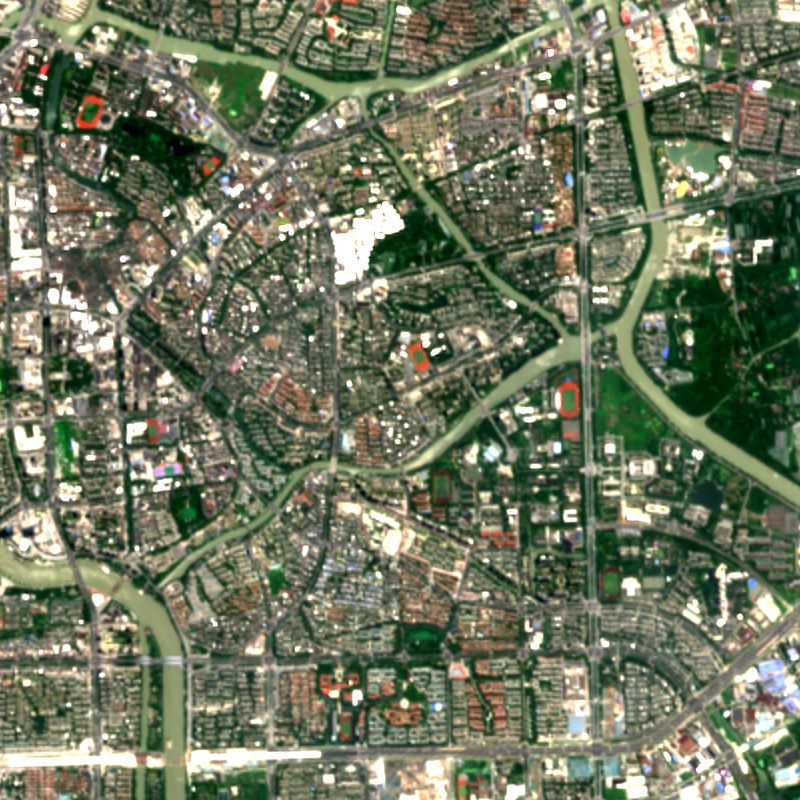}\\[-2pt]{\tiny Built-up}\phantomcaption\label{fig:s2-built-up}\end{subfigure}\hfill
        \begin{subfigure}[b]{0.13\textwidth}\centering\includegraphics[width=\linewidth]{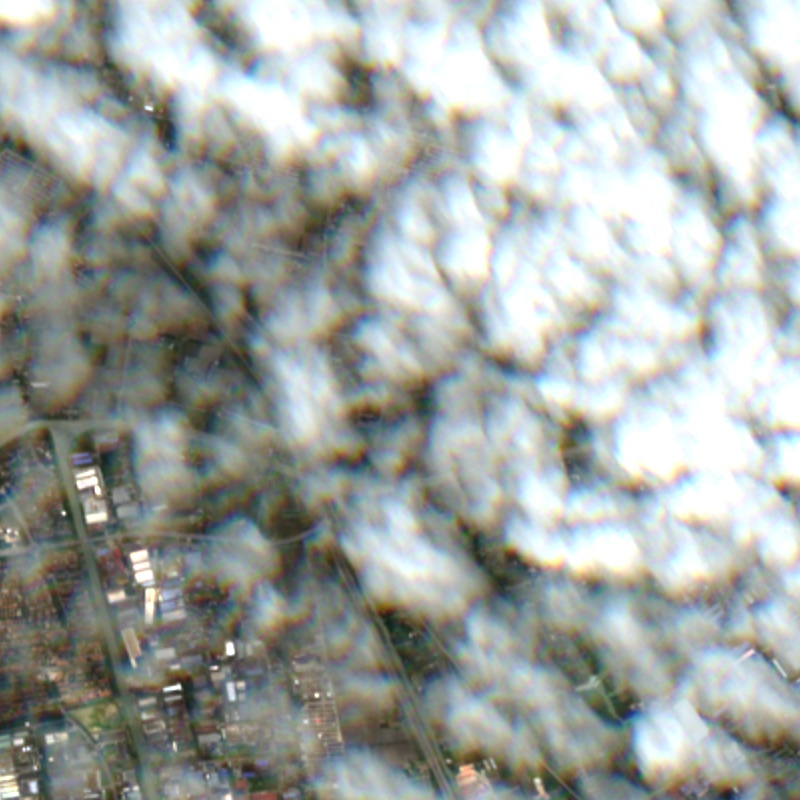}\\[-2pt]{\tiny Clouds}\end{subfigure}\hfill
        \begin{subfigure}[b]{0.13\textwidth}\centering\includegraphics[width=\linewidth]{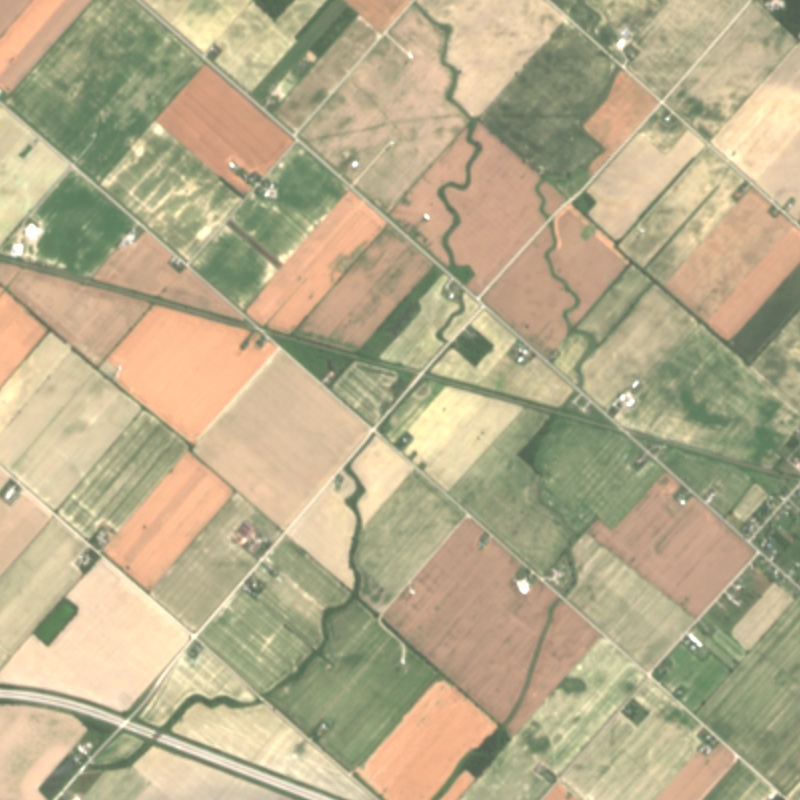}\\[-2pt]{\tiny Cropland}\phantomcaption\label{fig:s2-cropland}\end{subfigure}\hfill
        \begin{subfigure}[b]{0.13\textwidth}\centering\includegraphics[width=\linewidth]{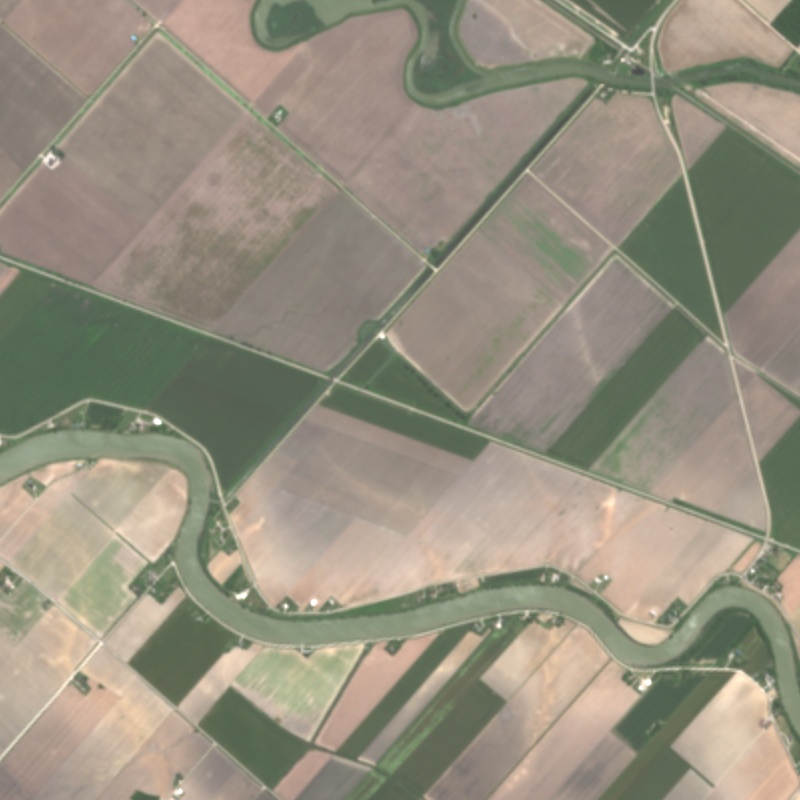}\\[-2pt]{\tiny Cropland}\end{subfigure}\hfill
        \begin{subfigure}[b]{0.13\textwidth}\centering\includegraphics[width=\linewidth]{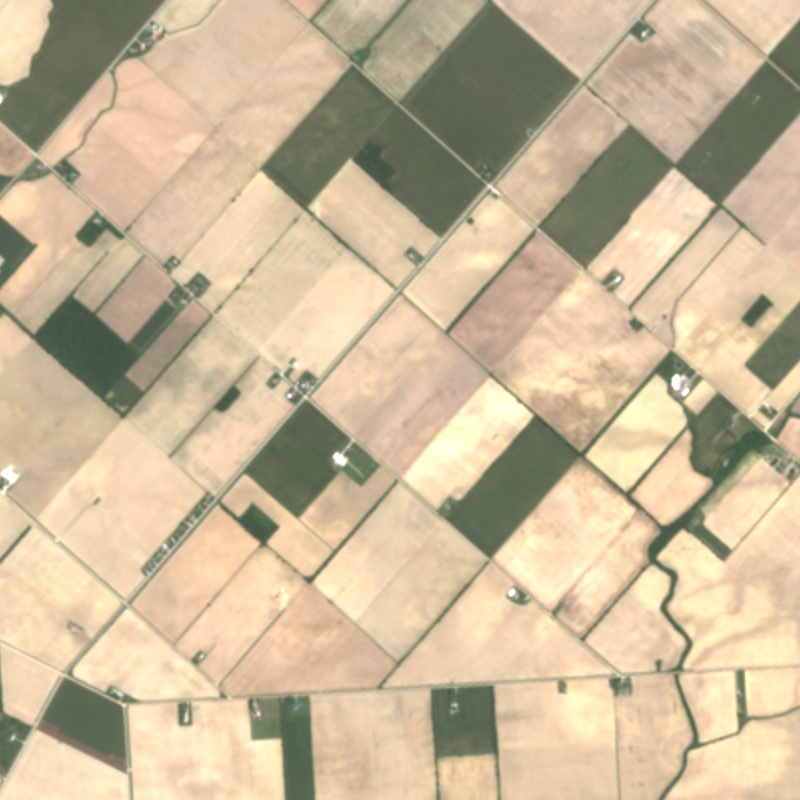}\\[-2pt]{\tiny Cropland}\end{subfigure}\hfill
        \begin{subfigure}[b]{0.13\textwidth}\centering\includegraphics[width=\linewidth]{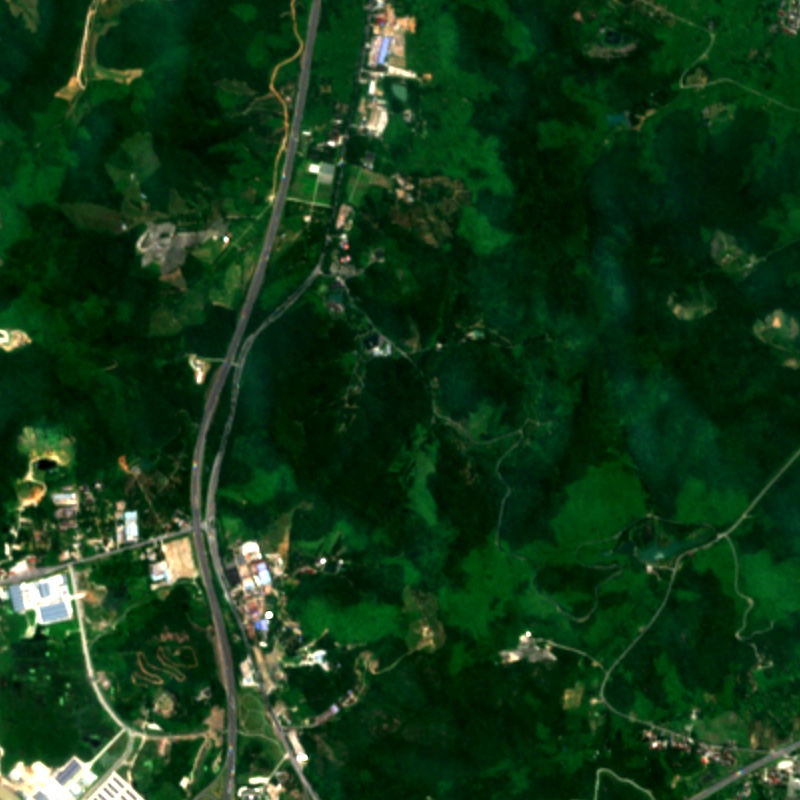}\\[-2pt]{\tiny Treecover}\phantomcaption\label{fig:s2-treecover}\end{subfigure}\hfill
        \begin{subfigure}[b]{0.13\textwidth}\centering\includegraphics[width=\linewidth]{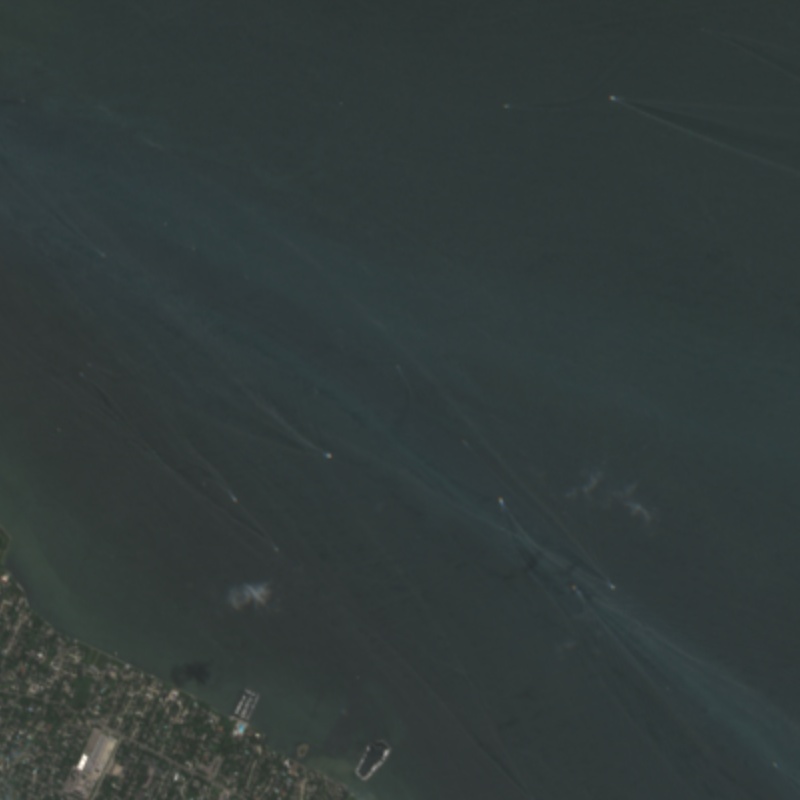}\\[-2pt]{\tiny Waterbody}\phantomcaption\label{fig:s2-waterbody}\end{subfigure}

        \vspace{4pt}
        %% -------- Group C: Cross Validation (5×1) --------
        \textbf{\scriptsize (c) Loft~\cite{loftorbital_imagery} --- Cross-Validation imagery}\\[2pt]
        \begin{subfigure}[b]{0.18\textwidth}\centering\includegraphics[width=\linewidth]{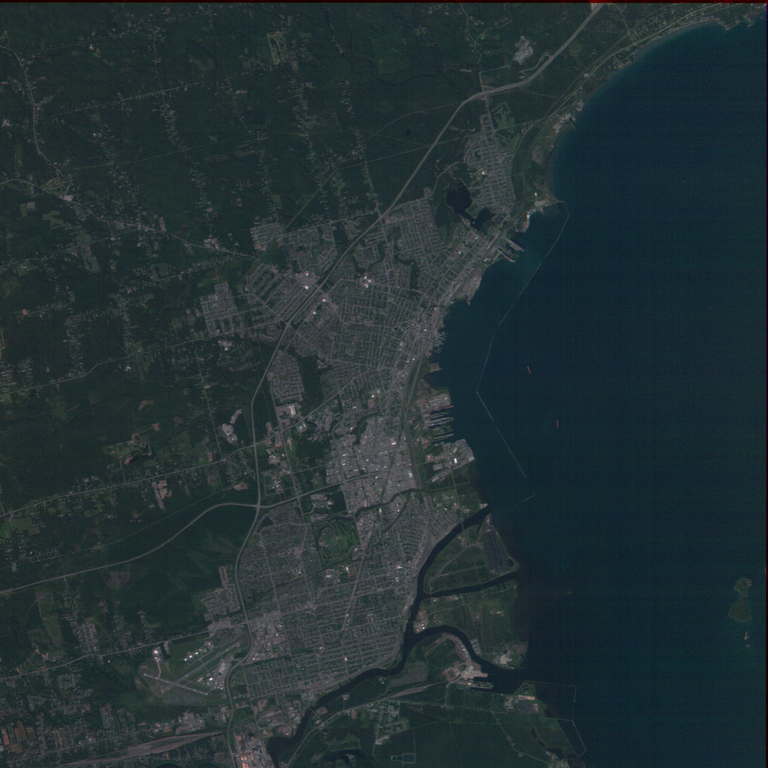}\\[-2pt]{\tiny 48.4°N, 89.2°W}\phantomcaption\label{fig:loft-coastal1}\end{subfigure}\hfill
        \begin{subfigure}[b]{0.18\textwidth}\centering\includegraphics[width=\linewidth]{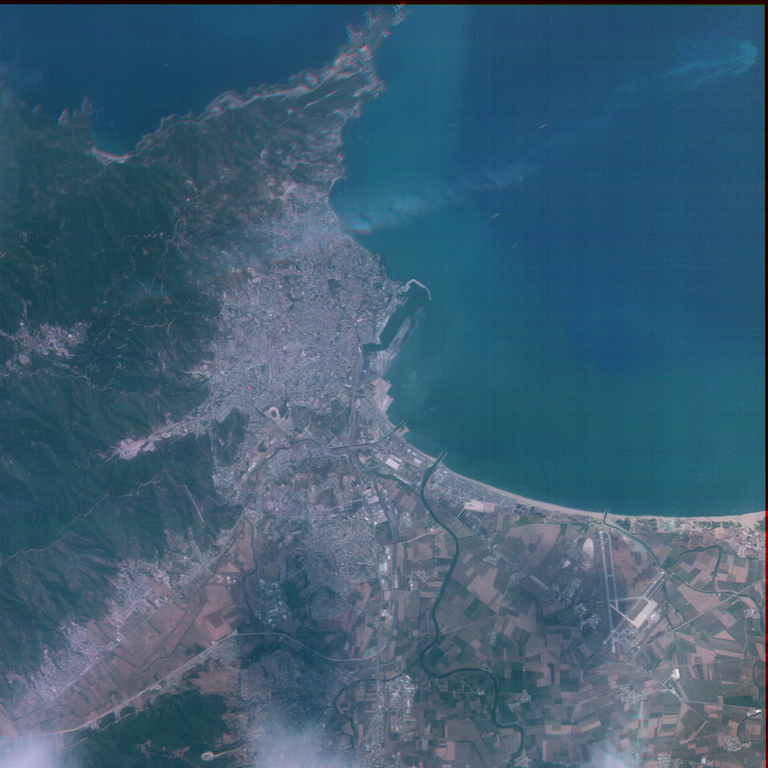}\\[-2pt]{\tiny 36.9°N, 7.8°E}\phantomcaption\label{fig:loft-coastal2}\end{subfigure}\hfill
        \begin{subfigure}[b]{0.18\textwidth}\centering\includegraphics[width=\linewidth]{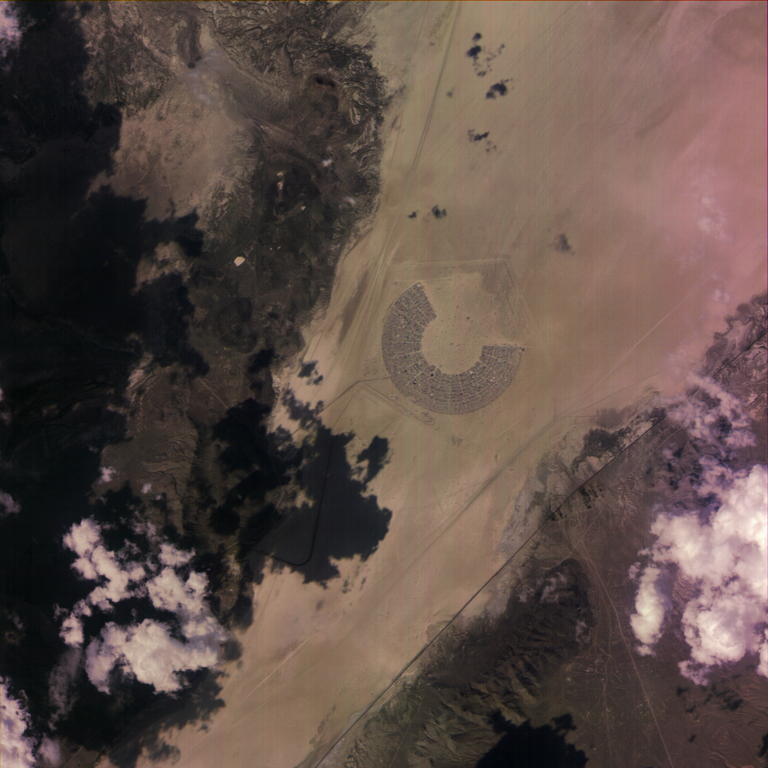}\\[-2pt]{\tiny 40.8°N, 119.2°W}\phantomcaption\label{fig:loft-desert}\end{subfigure}\hfill
        \begin{subfigure}[b]{0.18\textwidth}\centering\includegraphics[width=\linewidth]{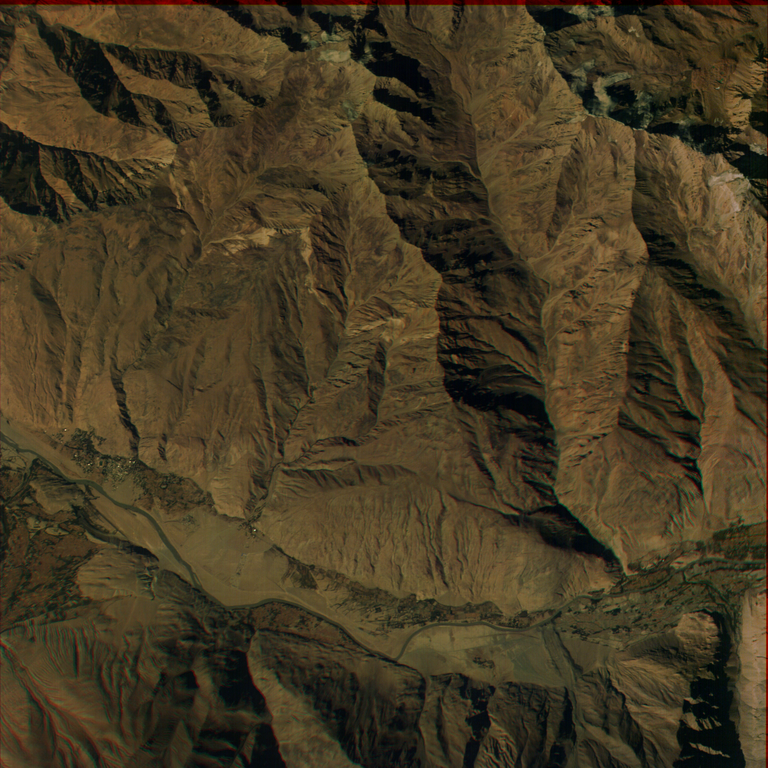}\\[-2pt]{\tiny 36.7°N, 71.7°E}\phantomcaption\label{fig:loft-mountain}\end{subfigure}\hfill
        \begin{subfigure}[b]{0.18\textwidth}\centering\includegraphics[width=\linewidth]{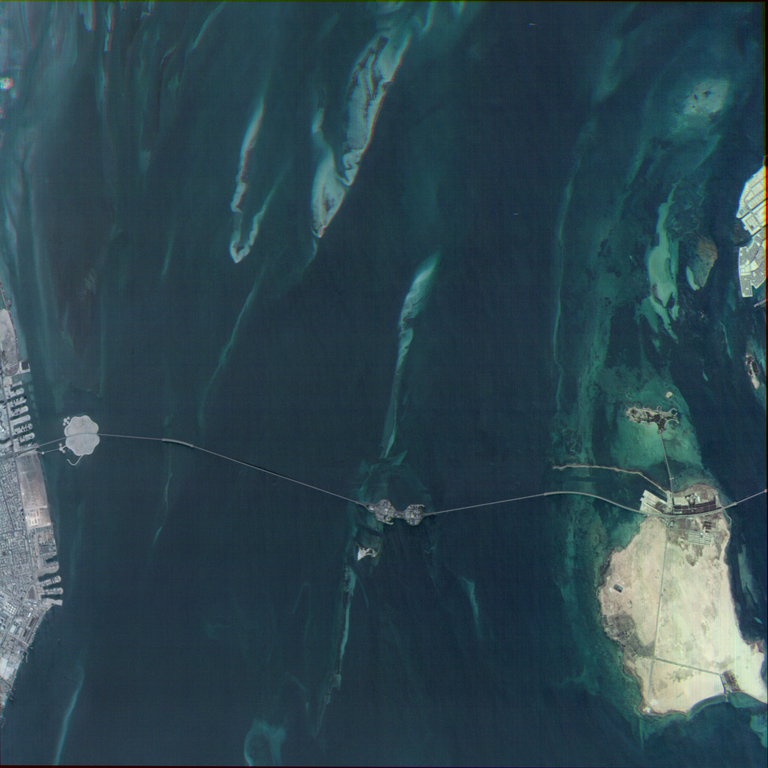}\\[-2pt]{\tiny 26.2°N, 50.3°E}\phantomcaption\label{fig:loft-bridge}\end{subfigure}

        \caption{Complete input imagery for the three evaluation datasets.
        (a)~Google AID: 18 classes curated from the 30-class benchmark.
        (b)~Sentinel-2: 7 tiles labeled via ESA WorldCover.
        (c)~Loft cross-validation: 5 images from imagers analogous to the YAM-9 camera. \textcopyright\ Loft Orbital 2026.}
        \label{fig:dataset_mosaic}
    \end{figure*}

    The curated dataset spans natural landscapes (Forest, Desert, Mountain, River, Beach) and anthropogenic
    infrastructure (Airport, Port, Industrial, Bridge, Highway Interchange, Parking, Storage Tanks, Railway Station,
    Commercial District, Venue, Residential, Water Reservoir, Agricultural). Class sizes range from 250 (Forest) to
    1{,}000 (Residential) with a mean of 442 images per class. Evaluation uses the identical deployment prompt and
    response parser described in Section~\ref{NAVIArchitecture}; no special evaluation harness is employed. The model
    configuration matches the flight build: Gemma~3 4B at 4-bit quantization via \texttt{bitsandbytes}. Inference was
    parallelized across 10 GPU workers for throughput during ground benchmarking.
    Figure~\ref{fig:dataset_mosaic} presents the complete input imagery for all three evaluation datasets.

    \subsubsection{Sentinel-2}
    \label{sec:datasetsentinel}
    The second dataset comprises manually curated Sentinel-2 Level-2A tiles sourced from the Copernicus
    Browser\footnote{\url{https://browser.dataspace.copernicus.eu/}}, rendered as true-color composites
    (bands B04, B03, B02) and exported as JPEG images at native 10\,m/pixel resolution.
    Ground truth labels were assigned by cross-referencing each tile's geographic coordinates against the
    ESA WorldCover 2021 product (v200)~\cite{zanaga2022esa}, a global 10\,m land-cover map derived from
    Sentinel-1 and Sentinel-2 data. The WorldCover 11-class taxonomy, Tree~cover, Shrubland, Grassland,
    Cropland, Built-up, Bare/sparse~vegetation, Snow~and~ice, Permanent~water~bodies, Herbaceous~wetland,
    Mangroves, and Moss~and~lichen, served as the label set for this experiment, augmented with an additional
    \textit{Clouds} class to evaluate the model's ability to flag unusable acquisitions, a critical operational
    requirement for autonomous onboard triage.
    The selected tiles span Built-up, Clouds, Cropland, Treecover, and Waterbody, chosen to represent visually
    distinct land-cover types at spatial scales comparable to operational LEO acquisitions.

    \subsubsection{Loft's Instrument Captures}
    \label{sec:datasetloft}
    The third dataset comprises imagery sourced from operational satellite imagers rather than public benchmarks.
    All images in this category were processed by the same NAVI-Orbital pipeline used for the
    preloaded datasets in the Flatsat or onboard YAM-9. Two subsets are distinguished by image origin:

    \paragraph{Cross-Validation Dataset}
    Five images captured by imagers materially similar to those onboard YAM-9 were uploaded to the Flatsat environment for pre-flight cross-validation of the onboard engine against never encountered scenes.  Additionally, these images were uploaded to the spacecraft
    prior to flight with one image from this set (Fig.~\ref{fig:loft-bridge}) processed as an in-orbit demonstration. Figure~\ref{fig:dataset_mosaic}(c) shows the full cross-validation set.

    \paragraph{YAM-9 Live}
    Two images captured directly by a YAM-9 imager during the mission were processed onboard in real-time and subsequently re-validated in Loft's ground test infrastructure. Captured images were 10-bit RGB prior to lossy conversion to 8-bit RGB. Detection outputs from these captures are presented in Section~\ref{sec:results-new-orbit-live}.

    \subsection{Metrics}
    Classification performance is assessed using overall accuracy, macro-averaged and weighted-average precision,
    recall, and F1-score. For each class $c$, let $\mathrm{TP}_c$, $\mathrm{FP}_c$, and $\mathrm{FN}_c$ denote
    true positives, false positives, and false negatives respectively:
    \begin{equation}
      \small
      P_c = \frac{\mathrm{TP}_c}{\mathrm{TP}_c + \mathrm{FP}_c}, \quad
      R_c = \frac{\mathrm{TP}_c}{\mathrm{TP}_c + \mathrm{FN}_c}, \quad
      F_{1,c} = \frac{2\,P_c\,R_c}{P_c + R_c}
    \end{equation}
    Macro-averaged scores are computed as the unweighted mean over all $C$ classes (e.g.\ $F_1^{\mathrm{macro}} = \frac{1}{C}\sum_c F_{1,c}$). A per-class confusion matrix captures inter-class misclassification patterns. Parse failure rate, the fraction of inferences in which the model's output could not be mapped to a valid label, serves as a proxy for hallucination rate. Average per-image inference latency is recorded to verify real-time feasibility under the spacecraft's processing budget.

    \subsection{System Configuration}
    \subsubsection{System Settings}
    Onboard image classification imposes a distinct set of constraints that directly shape each inference
    parameter. Execution follows a heterogeneous CPU--GPU pipeline:
    all transformer layers are offloaded to the GPU, which dominates the
    token-generation phase, while four CPU threads handle the surrounding
    orchestration work: tokenization, sampling, image I/O marshalling, and LangGraph state
    management. The context window is set to 20{,}000 tokens to accommodate high-resolution image patch embeddings alongside the full structured prompt without truncation, a requirement driven by the Pan-and-Scan visual integration strategy employed by the model~\cite{google2025gemma}. Output length
    is capped at 1{,}000 tokens to bound per-image latency and memory
    allocation within the spacecraft's processing budget. A sampling temperature of 0.2 suppresses
    stochastic variation in favor of deterministic, factual responses, which is critical for consistent
    regex-based label parsing across the retry loop. Finally, up to ten retries with a 30-second
    inter-attempt sleep provide resilience against the
    transient hardware and scheduling anomalies inherent to the space environment, ensuring autonomous
    throughput without requiring ground operator intervention during a pass.

    \subsubsection{Detection Prompt}
    Effective prompting of large language models requires precise specification of the task, the expected
    output format, and any constraints on model behavior~\cite{white2023prompt}. Key principles include
    assigning the model an explicit persona to anchor its reasoning context, providing a closed label set
    to eliminate open-ended hallucinations, issuing unambiguous output format directives to enable
    programmatic parsing, and keeping instructions concise so that attention is not diluted across
    extraneous tokens---all of which become especially important under the latency and retry budgets of an
    onboard autonomous system. The detection prompt template embodies these principles to instruct the VLM
    to classify an image into exactly one label from a mission-defined set and return a structured
    label--description pair:

    \begin{lstlisting}[style=prompt]
    You are an on-board Science Assistant. Analyze the image. Classify it into ONE of the following user defined labels.
    
    Labels: {user_label_list}
    Provide a factual description of the image content.
    
    Your answer MUST be in the format:
    'Label: <image_label>'
    'Description: <image_description>'
    \end{lstlisting}

    At runtime, \texttt{\{user\_label\_list\}} is populated from the experiment
    descriptor's label set (Section~\ref{sec:workflow}). The structured output
    format enables regex-based parsing within the Detector Agent's retry loop
    (Section~\ref{NAVIArchitecture}). The same prompt template is used across
    ground benchmarking, Flatsat validation, and in-orbit demonstration.

    \subsubsection{Dialogue Prompt}
    After detection completes, the Dialogue Agent uses the accumulated
    classification results as context for natural-language querying:

    \begin{lstlisting}[style=prompt]
    As Science Assistant, your task is to answer
    questions about image classification results.
    
    **CONTEXT**
    
    * **User defined labels: :**
    {user_label_list}
    
    *
    
    * **Full Classification Log (with timestamps):**
    {classification_results}
    
    ---
    
    **INSTRUCTIONS**
    Carefully use the context above to answer the
    user's question.
    
    **Question:** {user_input}
    \end{lstlisting}

    The \texttt{\{classification\_results\}} variable aggregates all detection
    outputs from the preceding phase, providing the model with the full
    operational context. Questions are sourced either from a pre-defined file
    (non-interactive mode) or from live operator input (interactive mode), as
    described in Section~\ref{sec:workflow}.

    \subsubsection{Sample Dialogue Questions}
    In non-interactive mode, the system processes a queue of pre-defined
    questions. The following is a representative subset from the flight
    configuration:

    \begin{lstlisting}[style=prompt]
    Find all descriptions that mention both a natural feature (like 'forest', 'river', 'vegetation') and a man-made structure (like 'building', 'road', 'highway'). List the source_image_path for each match.
    \end{lstlisting}

    The full question set comprises 16 queries spanning label enumeration,
    temporal analysis, cross-class comparison, and description-based spatial
    reasoning; additional prompt-response examples are shown in Table~\ref{tab:dialogue-showcase}.

\section{Testing, Validation and Deployment}
    The NAVI-Orbital test and validation lifecycle comprised five stages:
    \begin{enumerate}
        \item \textbf{On-ground testing} on representative hardware (Section~\ref{sec:results-AID});
        \item \textbf{Engineering model integration} on NVIDIA Jetson Orin development kits;
        \item \textbf{Flatsat validation} on the YAM-9 engineering ground segment (Section~\ref{sec:results-loft-flatsat});
        \item \textbf{Flight model deployment} on YAM-9 prelaunch and validation;
        \item \textbf{In-flight testing} on the YAM-9 after launch (Section~\ref{sec:results-new-orbit}).
    \end{enumerate}

    \subsection{On-ground Testing}
    \label{sec:on-ground-testing}
    During development, NAVI-Orbital was evaluated as both a Dockerized container and a bare-metal Python virtual environment. While the containerized approach was initially preferred for operational simplicity, it ultimately failed to provide reliable hardware acceleration on the target CPU and GPU due to library incompatibilities with Python 3.12 and CUDA. Consequently, the bare-metal Python virtual environment was selected as the primary deployment method for early testing. Despite this shift, the overarching operational workflow, including the file-based inbox/outbox contract that ensures cross-platform portability (detailed in Section~\ref{sec:workflow}), remained intact.

    Following the launch of YAM-9, the software ecosystem matured. In Q1 2026, NVIDIA released an updated Docker container that resolved these legacy dependency and hardware access limitations. Utilizing this new release, the authors successfully re-tested NAVI-Orbital within the ground Flatsat environment, finally achieving stable containerized execution with full hardware acceleration on the GPU.

    \subsection{Engineering Model Testing}
    %% Fig. 3 source moved up here so the float is queued before any V.B
    %% text is laid out, giving LaTeX a chance to place it at the top of
    %% the page where it is referenced.
    \begin{figure*}[!htb]
        \centering
        \begin{subfigure}[t]{0.48\textwidth}
            \centering
            \includegraphics[width=\linewidth]{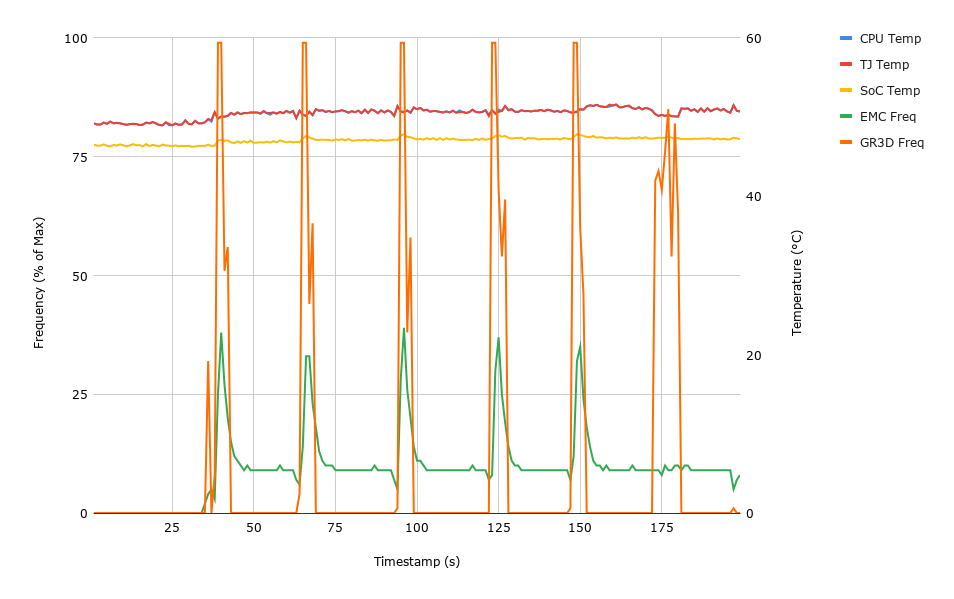}
            \caption{Frequency and Temperature}
        \end{subfigure}
        \hfill
        \begin{subfigure}[t]{0.48\textwidth}
            \centering
            \includegraphics[width=\linewidth]{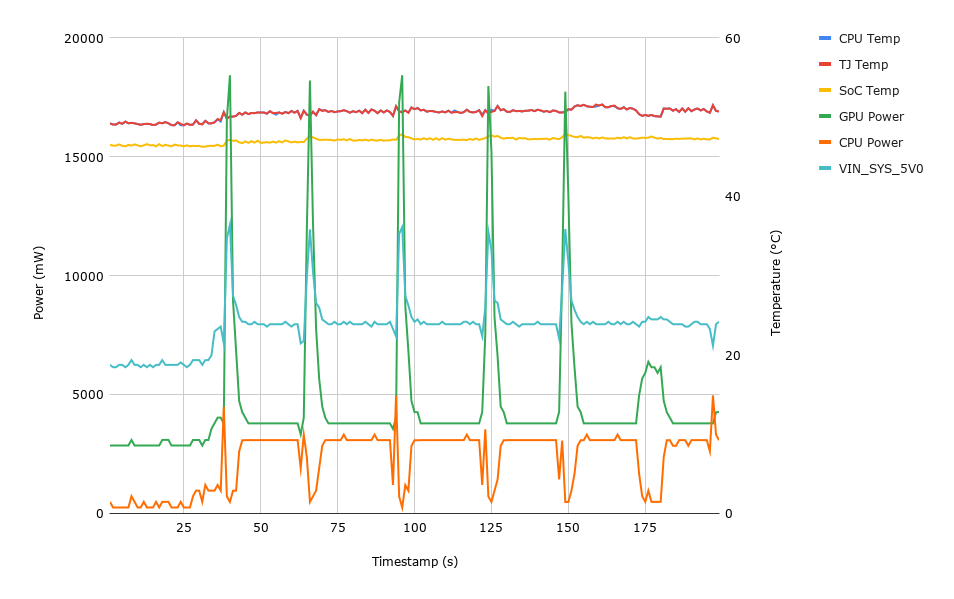}
            \caption{Power and Temperature}
        \end{subfigure}
        \caption{Comp2 Performance Benchmarks during the VLM stage. Left (a): Percent of maximum Frequency and Temperature. Right (b): Milliwatts and Temperature.}
        \label{fig:loft-benchmark}
    \end{figure*}

    The YAM-9 compute cluster comprises a heterogeneous mix of CPUs, GPUs, drivers, and operating systems, detailed in Table~\ref{tab:compute_cluster_loft}.

        \begin{table}[h]
            \centering
            \caption{YAM-9 Compute Cluster}
            \label{tab:compute_cluster_loft}
            \renewcommand{\arraystretch}{1.2}
            \footnotesize
            \begin{tabular}{llll}
                \toprule
                \textbf{Device} & \textbf{Specifications} & \textbf{Operating System}\\
                \midrule
                Comp0 & 4$\times$2\,GHz CPU0 & Ubuntu 20.04 LTS\\
                Comp1 & 12$\times$2\,GHz CPU1, 248\,TOPS GPU1 & Ubuntu 20.04 LTS\\
                Comp2 & 12$\times$2\,GHz CPU2, 248\,TOPS GPU2 & OpenEmbedded Scarthgap 5.0\\
                \bottomrule
            \end{tabular}
        \end{table}     

    To determine the optimal execution environment, the Engineering Models (EM) of the three computers were benchmarked using five images from the cross-validation dataset (Section~\ref{sec:datasetloft}). Comp2 was the only node supporting GPU acceleration, as the other nodes experienced library conflicts with NAVI-Orbital. As a baseline comparison, the ARM processor on Comp1 was used for CPU-only tests. 
    
    Table~\ref{tab:compute_cluster_loft_benchmark} summarizes execution times when processing the cross-validation set.

    \begin{table}[h]
        \centering
        \caption{NAVI-Orbital Performance on the Cross-Validation Dataset}
        \label{tab:compute_cluster_loft_benchmark}
        \renewcommand{\arraystretch}{1.2}
        \begin{tabular}{lll}
            \toprule
            \textbf{Device} & \textbf{Environment} & \textbf{Execution Time} \\
            \midrule
            Comp0 & Dockerized w/o HW Acceleration & 38~min \\
            Comp1 & Dockerized w/o HW Acceleration & 37~min \\
            Comp2 & Dockerized w/o HW Acceleration & 37~min \\
            Comp2 & Python venv w/ HW Acceleration & 13~min \\
            \bottomrule
        \end{tabular}
    \end{table}

    As expected, the bare-metal NAVI-Orbital deployment on Comp2 (utilizing both GPU2 and CPU2) significantly outperformed the other configurations. Specifically, total execution time on Comp2 for the cross-validation dataset was 13 minutes, compared to over 37 minutes for CPU-only configurations. The average VLM cycle time for the five processed images was 26.6 seconds.
    
    Figure~\ref{fig:loft-benchmark} illustrates CPU, GPU, and thermal utilization on Comp2 during VLM execution. During each of the five VLM cycles, the GPU Frequency (GR3D) maximizes the utilization of the external memory controller. Together, these metrics indicate extensive GPU utilization followed by elevated CPU processing (characterized by EMC Freq which quantifies external memory load). 
    
    Additionally, power telemetry aligns with this operational profile, showing spikes in GPU energy consumption followed by sustained CPU load.  Metrics plotted are~\cite{nvidia2026tegrastats}~\cite{nvidia2026orinpower}:
    \begin{itemize}
        \item CPU Temp: CPU core temperature block
        \item TJ Temp: Core junction temperature (critical to GPU throttling)
        \item SoC Temp: First of three SoC measurement points
        \item EMC Freq: Percentage of EMC memory bandwidth, relative to maximum set point
        \item GR3D Freq: Proportion of GPU activation time in a period
        \item GPU Power: Power consumed by the GPU subsystem
        \item CPU Power: Power consumed by the CPUs
        \item {VIN\_SYS\_5V0}: Power consumed by the external memory interfaces
    \end{itemize}

    At the conclusion of Engineering Model testing, the software configuration was frozen in preparation for the Flatsat campaign and the subsequent flight upload to YAM-9. This freeze was the consequence of several rapidly evolving technologies (models, frameworks, containerizations) not converging at the time of this deployment, which forced down-selecting to a user space deployment. Without the modularity of a containerized approach, deploying directly to user space severely restricted upgrade flexibility; any subsequent model changes would have required repeating the entire validation chain, from EM benchmarking through Flatsat. Newer Gemma releases and alternative architectures evaluated after this point were deferred from the flight build.

    \subsection{Flatsat Validation}
    \label{sec:results-loft-flatsat}
    Loft's Flatsat provides a flight-representative staging area to integrate hardware, software, and data for validation. For NAVI-Orbital, the Flatsat environment facilitated an ecosystem for operators to uncover integration challenges at the satellite level prior to flight. Unlike the development environments described above, the Flatsat operates as a true staging area for flight hardware.
    
    Datasets and findings from the Flatsat campaign are presented in Section~\ref{sec:results-new-engflatsat}; the validated configuration was uploaded and executed on the Flight Hardware.

\section{Results}
\label{sec:results}

\subsection{Order of Events}
\label{sec:order_of_events}

\begin{figure*}[!htb]
    \centering
    \includegraphics[width=\textwidth]{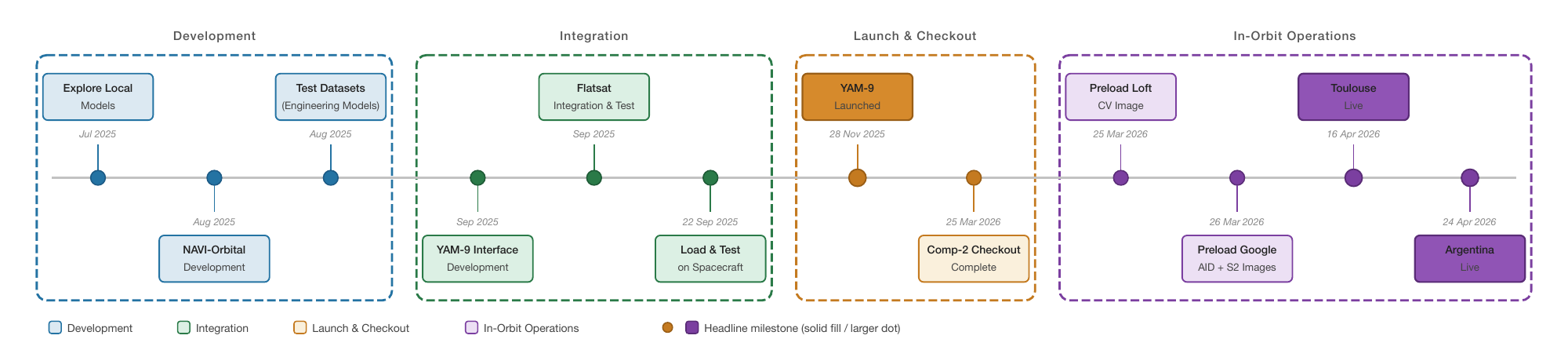}
    \caption{NAVI-Orbital development timeline. Twelve milestones spanning ten months, grouped into four phases from initial software exploration through live in-orbit demonstrations.}
    \label{fig:timeline}
\end{figure*}

The NAVI-Orbital work spanned ten months, from initial exploration of local vision-language models in July 2025 through live in-orbit captures over Toulouse and Argentina in April 2026. Figure~\ref{fig:timeline} summarizes the twelve principal milestones, grouped into four phases.

The Development phase characterized candidate models against engineering datasets and assembled the agent graph described in Section~\ref{NAVIArchitecture}; quantitative outcomes from this phase are reported in Section~\ref{sec:ground_benchmarking}. The Integration phase moved the validated stack from a flight-representative computer at JPL onto Loft Orbital's YAM-9 Flatsat and finally onto the flight vehicle, with results reported in Section~\ref{sec:results-new-engflatsat}. Launch occurred on 28 November 2025, followed by a four-month spacecraft checkout. The In-Orbit Operations phase then preloaded reference imagery onto the onboard compute element and executed two live capture campaigns, detailed in Section~\ref{sec:results-new-orbit}.

% ------------------------------------------------------------------ (A)
\subsection{Ground Benchmarking on Google AID}
\label{sec:results-AID}
\label{sec:ground_benchmarking}
The ground benchmarking experiment evaluated Gemma~3 4B on the full 7{,}960-image curated AID
dataset. Of these, 7{,}956 images produced valid classifications, yielding an overall accuracy of
\textbf{88.16\%} and a macro-averaged F1-score of \textbf{0.87}.

\begin{figure}[!htb]
    \centering
    \includegraphics[width=\columnwidth]{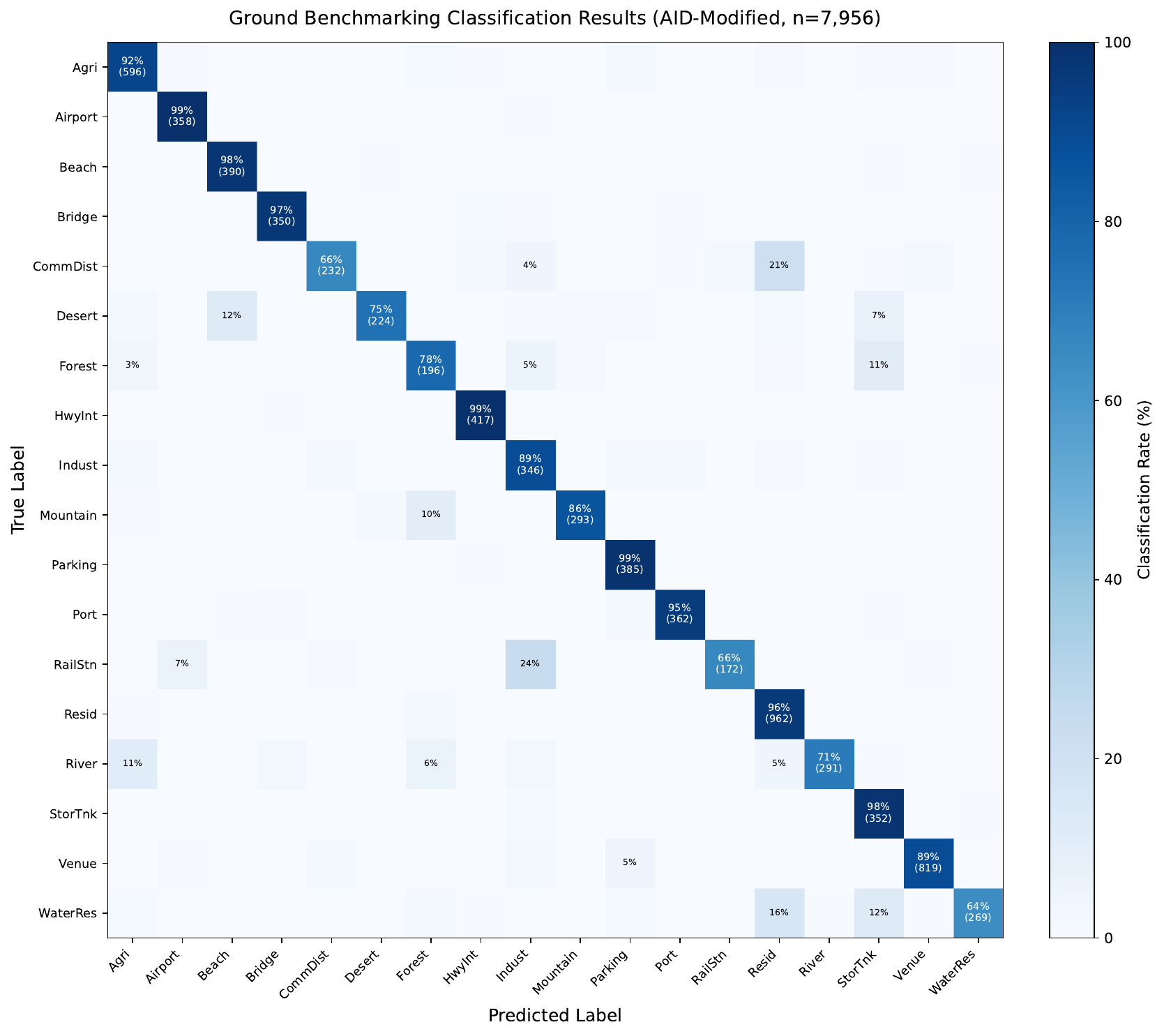}
    \caption{Confusion matrix for 18-class zero-shot classification on the curated AID dataset (7{,}956 valid
    predictions). Diagonal dominance confirms strong per-class discrimination; off-diagonal mass concentrates in
    semantically plausible confusions (e.g., Commercial District~$\rightarrow$~Residential,
    Railway Station~$\rightarrow$~Industrial).}
    \label{fig:confusion_matrix}
\end{figure}

\begin{table}[!htb]
\centering
\caption{Per-class classification metrics on the curated AID dataset (7{,}956 valid predictions). The model
achieves 88.16\% overall accuracy with a macro-averaged F1-score of 0.87.}
\label{tab:per_class_metrics}
\footnotesize
\begin{tabular}{lcccc}
\toprule
\textbf{Class} & \textbf{Prec.} & \textbf{Rec.} & \textbf{F1} & \textbf{Supp.} \\
\midrule
Agricultural & 0.88 & 0.92 & 0.90 & 647 \\
Airport & 0.92 & 0.99 & 0.95 & 360 \\
Beach & 0.91 & 0.97 & 0.94 & 400 \\
Bridge & 0.94 & 0.97 & 0.96 & 360 \\
Commercial District & 0.88 & 0.66 & 0.76 & 350 \\
Desert & 0.96 & 0.75 & 0.84 & 300 \\
Forest & 0.66 & 0.78 & 0.72 & 250 \\
Highway Interchange & 0.95 & 0.99 & 0.97 & 420 \\
Industrial & 0.72 & 0.89 & 0.80 & 390 \\
Mountain & 0.98 & 0.86 & 0.92 & 340 \\
Parking & 0.81 & 0.99 & 0.89 & 390 \\
Port & 0.95 & 0.95 & 0.95 & 380 \\
Railway Station & 0.93 & 0.66 & 0.77 & 260 \\
Residential & 0.84 & 0.96 & 0.90 & 1{,}000 \\
River & 0.98 & 0.71 & 0.82 & 410 \\
Storage Tanks & 0.74 & 0.98 & 0.84 & 360 \\
Venue & 0.98 & 0.89 & 0.93 & 920 \\
Water Reservoir & 0.97 & 0.64 & 0.77 & 419 \\
\midrule
\textbf{Macro avg} & \textbf{0.89} & \textbf{0.87} & \textbf{0.87} & 7{,}956 \\
\textbf{Weighted avg} & \textbf{0.89} & \textbf{0.88} & \textbf{0.88} & 7{,}956 \\
\bottomrule
\end{tabular}
\end{table}

Figure~\ref{fig:confusion_matrix} and Table~\ref{tab:per_class_metrics} present the full confusion matrix and
per-class metrics, respectively. The strongest performers are Highway Interchange (F1\,=\,0.97), Bridge (0.96),
and Airport (0.95), structurally distinctive scenes with unambiguous visual signatures at aerial resolution.
The weakest classes are Forest (F1\,=\,0.72), Commercial District (0.76), and Water Reservoir (0.77), each
exhibiting semantic overlap with neighboring categories. The dominant confusion patterns are Commercial
District~$\rightarrow$~Residential (21\% of Commercial District images), Water
Reservoir~$\rightarrow$~Residential and Storage Tanks, and Railway Station~$\rightarrow$~Industrial (24\%).
These misclassifications are semantically plausible: aerial views of commercial districts and dense residential
areas share rooftop and road-network features, while water reservoirs adjacent to built-up areas can resemble
storage infrastructure. The parse failure rate was 4 out of 7{,}960 images (0.05\%), indicating reliable prompt
adherence; a detailed analysis is provided in Section~\ref{sec:hallucination}. Average per-image inference time was
5.6\,s on a desktop GPU.

These results establish a performance baseline for the deployment-identical prompt and parser, against which
Flatsat and in-orbit predictions are compared in Section~\ref{sec:results-loft-flatsat} and Section~\ref{sec:results-new-orbit}.
% ------------------------------------------------------------------ (B)
\subsection{Engineering Model and Flatsat Validation}
\label{sec:results-new-engflatsat}

Prior to flight, preloaded experiments were validated on the engineering
model and the Flatsat environment (see Section~\ref{sec:results-loft-flatsat} for compute-cluster benchmarking and Fig.~\ref{fig:loft-benchmark}).

% - - - - - - - - - - - - - - - - - - - - - - - - - - - (B.i)
\subsubsection{Preloaded Experiments}
\label{sec:results-new-preloaded}

%% Table V (dialogue showcase) source moved up here so the float is queued
%% before any text in this subsection is laid out, letting LaTeX place it
%% at the top of the page where it is referenced.
\begin{table*}[!htb]
    \centering
    \footnotesize
    \begin{tabular}{p{0.28\linewidth} p{0.68\linewidth}}
        \toprule
        \textbf{Question} & \textbf{Model Response} \\
        \midrule
        % --- 1. Semantic cross-reference (Google AID) ---
        Based on the descriptions, identify all images that mention a water feature (e.g., ocean, river, harbor). List the source image path and its primary label.
        &
        Beach (bc\_1.jpg), Bridge (bd\_4.jpg), River (rv\_6.jpg), Water Reservoir (wr\_59.jpg), Port (pt\_3.jpg). \\
        \midrule
        % --- 2. Operational occlusion detection (Sentinel-2) ---
        Was there any image where the view of the surface was significantly occluded by clouds?
        &
        Yes, one image (cl.jpg) was classified as Clouds. \\
        \midrule
        % --- 3. Comparative scene characterization (Google AID) ---
        Compare the primary characteristics of any `Residential' areas versus any `Commercial' areas noted in the results.
        &
        The Residential area is described as a densely populated area with numerous houses, closely spaced, and interspersed with trees. The Commercial area is characterized by storage tanks, buildings, parking areas, and roadways. The primary difference is housing and green spaces vs.\ industrial facilities and infrastructure. \\
        \midrule
        % --- 4. Infrastructure-proximity reasoning (Google AID) ---
        Which industrial or transport-related classifications are explicitly described as being located near residential areas?
        &
        The Railway Station and Storage Tanks are explicitly described as being located near residential areas. The Railway Station description states: ``The surrounding area appears to be a busy commercial and residential area.'' \\
        \bottomrule
    \end{tabular}
    \caption{Four selected dialogue exchanges from the preloaded experiments
    demonstrating semantic-understanding capabilities of the Dialogue Agent:
    cross-image semantic search, operational occlusion detection, comparative
    scene characterization, and infrastructure-proximity reasoning.  Questions 1,
    3, and 4 are from Google AID; question 2 is from Sentinel-2.  In every case
    the agent answers from the textual descriptions of the scenes, not from the
    class labels.}
    \label{tab:dialogue-showcase}
\end{table*}

Prior to flight, two preloaded experiment suites were executed through
the deployment-identical NAVI-Orbital pipeline to verify end-to-end
functionality, from image ingestion and VLM inference through JSON
serialization and dialogue interrogation.  Unlike the ground benchmarking
in Section~\ref{sec:results-AID}, which evaluates statistical accuracy over
7{,}960 images, these experiments constitute a \emph{functional verification}:
one representative image per class for Google AID and a curated subset
of Sentinel-2 land-cover tiles, confirming that every pipeline stage
produces correct output on representative inputs.

Table~\ref{tab:preloaded-summary} summarizes detection accuracy.
Both experiments achieved 100\% classification accuracy, with every image
correctly assigned to its ground-truth label on the first inference
attempt (no retries required). The Google AID experiment covered all 18
user-defined classes, while the Sentinel-2 experiment detected 5 of the
12 ESA WorldCover classes present in the 7-tile subset.

\begin{table}[h]
    \centering
    \caption{Preloaded experiment detection summary.}
    \label{tab:preloaded-summary}
    \footnotesize
    \begin{tabular}{lccc}
        \toprule
        \textbf{Dataset} & \textbf{Images} & \textbf{Correct} & \textbf{Accuracy} \\
        \midrule
        Google AID (18 classes) & 18 & 18 & 100\% \\
        Sentinel-2 (12 classes) &  7 &  7 & 100\% \\
        \midrule
        \textbf{Total} & \textbf{25} & \textbf{25} & \textbf{100\%} \\
        \bottomrule
    \end{tabular}
\end{table}

After detection, the Dialogue Agent was exercised with a battery of
16 structured questions per experiment, probing the VLM's ability to
reason \emph{semantically} over the accumulated classification log rather
than merely retrieve labels.  Table~\ref{tab:dialogue-showcase} shows four
representative exchanges: cross-image semantic search (water features),
operationally relevant occlusion detection (clouds), comparative scene
characterization (residential vs.\ commercial), and infrastructure-proximity
reasoning.  In each case the agent reads the natural-language descriptions
produced by the detector and answers questions about \emph{content} rather
than \emph{category}.  This is the capability that makes the Dialogue Agent
useful for downlink prioritization: an operator can ask ``is there evidence
of cloud cover?'' or ``which scenes contain human infrastructure?'' without
having to design a new classifier.

% - - - - - - - - - - - - - - - - - - - - - - - - - - - (B.ii)
\subsubsection{Cross-Validation on Flatsat}
\label{sec:results-new-cv}

Five cross-validation images were processed on the Flatsat to verify end-to-end system behavior; detection outputs are shown in Table~\ref{tab:cv-detections}.
\begin{table*}[!htb]
    \centering
    \begin{tabular}{@{}m{0.16\textwidth}@{\hspace{8pt}}p{0.80\textwidth}@{}}
        \toprule
        \textbf{Image} & \textbf{Detection} \\
        \midrule
        \includegraphics[width=0.16\textwidth]{images/loft-cross-validation/session_data_1756241154-rgb.png} & \footnotesize
        The image shows a densely populated urban area with a significant concentration of buildings and roads. The city is situated along a coastline, with a large body of water to the east. There are numerous industrial and port facilities visible, suggesting a strong commercial and transportation hub. The layout of the city indicates a well-developed infrastructure with a grid-like street pattern. There are also some residential areas interspersed within the commercial zone. \\[4pt]
        \includegraphics[width=0.16\textwidth]{images/loft-cross-validation/session_data_1756303494-rgb.png} & \footnotesize
        The image shows a large urban area, likely a city, situated along a coastline. The city is densely populated with buildings, suggesting a commercial district. There are also areas of farmland and agricultural land visible inland, and a large body of water (likely a sea or ocean) along the coast. The presence of a port and industrial areas indicates a significant commercial activity. The image also shows a highway interchange and a railway station. The surrounding terrain consists of mountains and hills. \\[4pt]
        \includegraphics[width=0.16\textwidth]{images/loft-cross-validation/session_data_1756506847-rgb.png} & \footnotesize
        The image shows an aerial view of a vast, arid desert landscape. A large, circular structure resembling a parking lot or a temporary installation is prominently featured in the center of the frame. The surrounding terrain is characterized by rolling sand dunes and rocky outcrops. Scattered small structures and patches of vegetation are visible in the lower right corner. Dark storm clouds are present in the upper left corner of the image, partially obscuring the view. \\[4pt]
        \includegraphics[width=0.16\textwidth]{images/loft-cross-validation/session_data_1757670720-rgb.png} & \footnotesize
        The image depicts a rugged, mountainous landscape. Steep, layered cliffs and canyons dominate the scene, exhibiting significant erosion patterns. A small settlement is nestled within a valley at the base of the mountains, suggesting a human presence in a challenging terrain. The color palette is primarily earthy tones, with browns, tans, and grays, indicative of rock and soil. \\[4pt]
        \includegraphics[width=0.16\textwidth]{images/loft-cross-validation/session_data_1757761313-rgb.png} & \footnotesize
        The image shows an aerial view of a port facility located in a coastal area. A long, dark bridge extends out into the water, connecting the port to a small island. The port itself is comprised of numerous docks, piers, and buildings, suggesting a busy commercial activity. The surrounding water is deep blue, with some areas showing coral reefs and rocky formations. There are also several smaller islands and land masses visible near the port. \\
        \bottomrule
    \end{tabular}
    \caption{Cross-validation detection results from the Flatsat environment.
    Each image was processed end-to-end through the NAVI-Orbital pipeline. \textcopyright\ Loft Orbital 2026.}
    \label{tab:cv-detections}
\end{table*}

Critically, these five images were never encountered during development
or prompt tuning: they originate from imagers materially analogous to the
YAM-9 camera, acquired during a previous mission, and were deliberately
withheld to serve as an unseen validation set.  The pipeline's correct
classification of all five scenes (Table~\ref{tab:cv-detections})
therefore demonstrates zero-shot generalization. The VLM and
prompt template produce accurate descriptions on imagery that
the system was not specifically optimized for, captured by hardware
representative of the flight instrument.  This result is significant
because it establishes that NAVI-Orbital's classification capability
extends beyond curated benchmark datasets to operationally realistic
imagery, a prerequisite for autonomous onboard inference.

% ------------------------------------------------------------------ (C)
\subsection{In-Orbit Results}
\label{sec:results-new-orbit}

Following the YAM-9 launch, commissioning, and transition to operations,
the team tasked the satellite to process a subset of preloaded and
cross-validation imagery and later to capture and process live Earth
observations.

% - - - - - - - - - - - - - - - - - - - - - - - - - - - (C.i)
\subsubsection{Preloaded and Cross-Validation Post-Processing}
\label{sec:results-new-orbit-preload}
\label{sec:results-new-orbit-cv}

A subset of preloaded images from the AID, Sentinel-2, and cross-validation datasets were processed onboard following commissioning to confirm satellite and GPU2 health and to verify match-rate consistency between ground and orbital inference. Detection outputs are shown in Table~\ref{fig:preloaded-samples}; they were consistent with pre-launch benchmarks and matched the corresponding Flatsat results.

\begin{table*}[!htb]
    \centering
    \begin{tabular}{@{}m{0.16\textwidth}@{\hspace{8pt}}p{0.80\textwidth}@{}}
        \toprule
        \textbf{Image} & \textbf{Detection} \\
        \midrule
        \includegraphics[width=0.16\textwidth]{images/preloaded-aid/al_4.jpg} & \footnotesize
        \textit{Preloaded (AID), Agricultural.}
        The image shows a rural landscape dominated by agricultural fields. The land is divided into numerous rectangular plots, suggesting a systematic farming operation. The fields vary in color, indicating different crops or stages of growth. There are also patches of forest interspersed among the fields, and a road running through the center of the image. \\[4pt]
        \includegraphics[width=0.16\textwidth]{images/preloaded-s2/bu.jpg} & \footnotesize
        \textit{Preloaded (Sentinel-2), Built-up.}
        The image shows a densely urbanized area with a complex network of streets, buildings, and canals. There is a significant amount of built-up infrastructure, including residential and commercial buildings, and a network of waterways running through the city. The color palette indicates a mix of materials, including rooftops, roads, and possibly some vegetation patches within the urban environment. The overall impression is of a highly developed metropolitan area. \\[4pt]
        \includegraphics[width=0.16\textwidth]{images/loft-cross-validation/session_data_1757761313-rgb.png} & \footnotesize
        \textit{Cross-Validation (Loft), 26.2°N, 50.3°E.}
        The image shows an aerial view of a port facility located in a coastal area. A long, dark bridge extends out into the water, connecting the port to a small island. The port itself is comprised of numerous docks, piers, and buildings, suggesting a busy commercial activity. The surrounding water is deep blue, with some areas showing coral reefs and rocky formations. There are also several smaller islands and land masses visible near the port. \\
        \bottomrule
    \end{tabular}
    \caption{Representative detection outputs processed onboard YAM-9: one preloaded AID image (Agricultural), one preloaded Sentinel-2 tile (Built-up), and one cross-validation Loft image (26.2°N, 50.3°E). Outputs match the corresponding Flatsat ground-test results, confirming consistent inference between ground and orbit. \textcopyright\ Loft Orbital 2026.}
    \label{fig:preloaded-samples}
\end{table*}

% - - - - - - - - - - - - - - - - - - - - - - - - - - - (C.iii)
\subsubsection{Live}
\label{sec:results-new-orbit-live}

\begin{figure*}[!htb]
    \centering
    \begin{minipage}[c]{0.45\textwidth}
        \vspace{0pt}\centering
        \includegraphics[width=\linewidth]{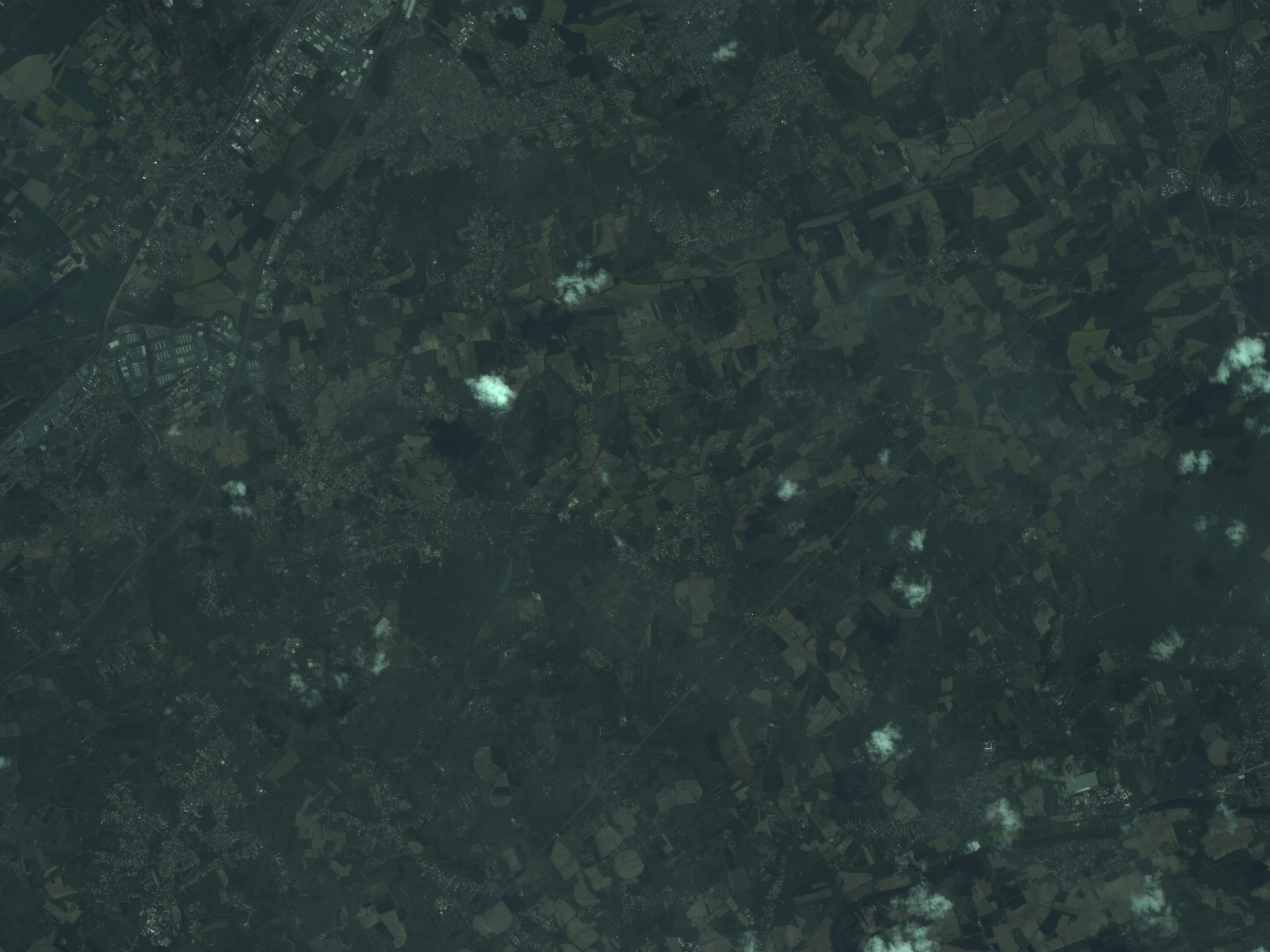}
    \end{minipage}\hfill
    \begin{minipage}[c]{0.50\textwidth}
        \vspace{0pt}\footnotesize
        \textbf{Detection} \par\smallskip\hrule\smallskip
        \textit{Label:} Residential \par\smallskip
        The image shows a nighttime satellite view of a densely populated area.
        Numerous residential areas are visible, characterized by a high
        concentration of light sources, indicating buildings and homes.
        The areas are interspersed with patches of darker areas, likely
        representing parks, green spaces, or undeveloped land. There are also
        some areas with more concentrated light, possibly indicating commercial
        or industrial zones within the residential area.

        \bigskip
        \textbf{Dialogue (2 of 17 scripted questions)} \par\smallskip\hrule\smallskip
        \textit{Q:} Identify any classifications where a natural environment is described as being adjacent to or integrated with a human-developed area. \par\smallskip
        \textit{A:} The first classification describes residential areas interspersed with parks, green spaces, or forested areas. The second classification describes residential areas with scattered parks or undeveloped land. \par\medskip
        \textit{Q:} For any image described as a railway hub, what other types of infrastructure (like roads or storage facilities) are there? \par\smallskip
        \textit{A:} The images described as Residential show roads and buildings, parks and green spaces, and residential areas with distinct patterns of development. However, the context does not provide any information about railway hubs or the infrastructure associated with them. \par\medskip
    \end{minipage}
    \caption{Live Earth observation~1: Toulouse, France (43.76°N, 1.38°E).
    Uncorrected 10-bit$\rightarrow$8-bit capture processed autonomously onboard YAM-9. \textcopyright\ Loft Orbital 2026.}
    \label{fig:hero-toulouse}
    \label{fig:loft-raw-tol}
\end{figure*}

\begin{figure*}[!htb]
    \centering
    \begin{minipage}[c]{0.45\textwidth}
        \vspace{0pt}\centering
        \includegraphics[width=\linewidth]{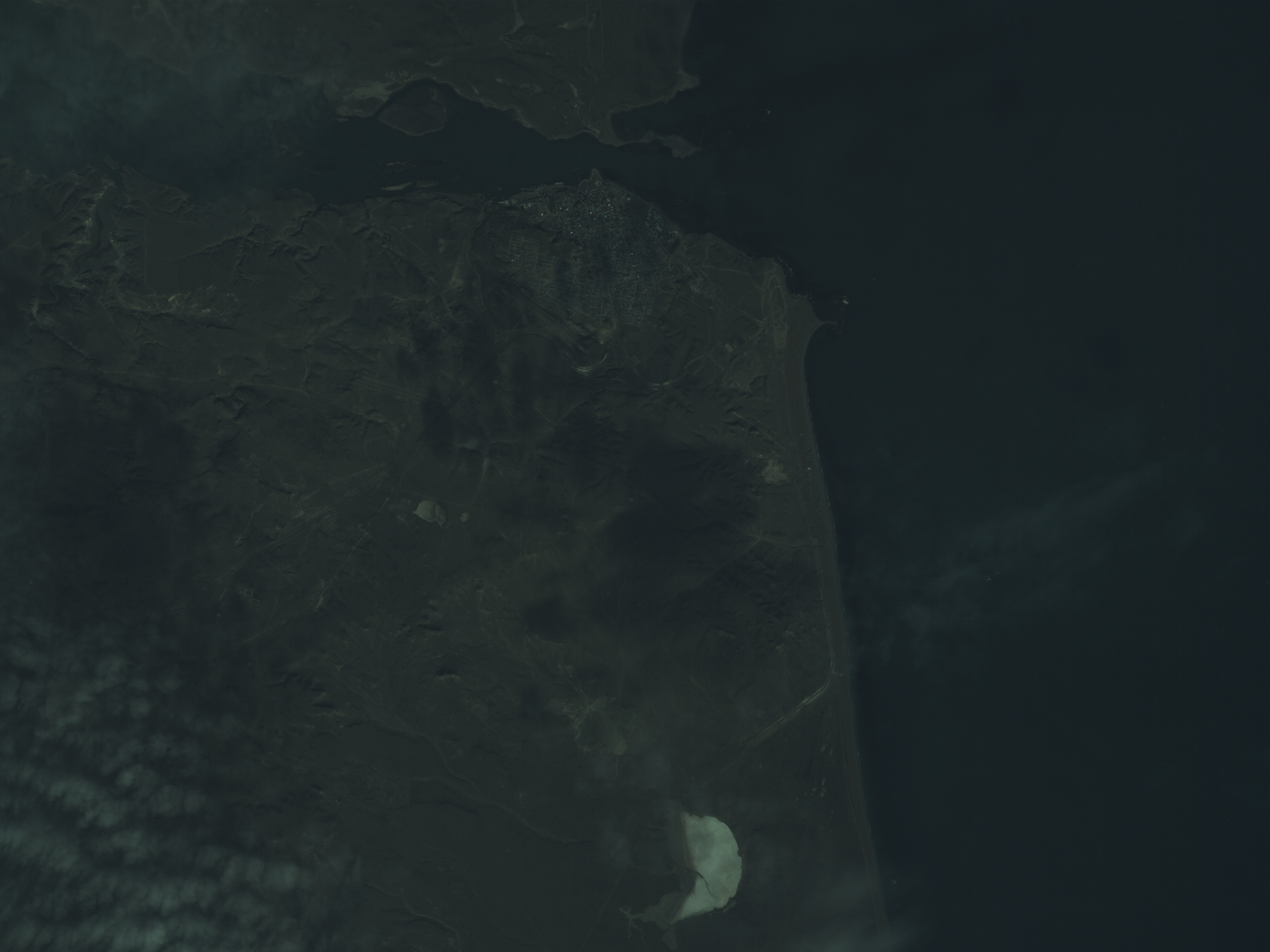}
    \end{minipage}\hfill
    \begin{minipage}[c]{0.50\textwidth}
        \vspace{0pt}\footnotesize
        \textbf{Detection} \par\smallskip\hrule\smallskip
        \textit{Label:} Beach \par\smallskip
        The image depicts a coastal area at night. The dominant feature is a
        dark, sandy beach extending along the shoreline. A body of water,
        likely the ocean, is visible on the right side of the image, and a
        small, bright area, possibly a city or town, is situated on a peninsula
        extending into the water. The overall tone is dark, with a slight glow
        emanating from the distant lights of the city.

        \bigskip
        \textbf{Dialogue (3 of 17 scripted questions)} \par\smallskip\hrule\smallskip
        \textit{Q:} Based on the descriptions, identify all images that mention a water feature (e.g., ocean, river, harbor). List the source image path and its primary label. \par\smallskip
        \textit{A:} scene\_0\_band\_0.png (Beach). \par\medskip
        \textit{Q:} Based on the descriptions, compare the primary characteristics of any `Residential' areas versus any `Commercial' areas noted in the results. \par\smallskip
        \textit{A:} The provided classification log does not contain any `Residential' or `Commercial' areas. It only identifies a `Beach' area. \par\medskip
        \textit{Q:} Identify any classifications where a natural environment is described as being adjacent to or integrated with a human-developed area. \par\smallskip
        \textit{A:} The image classification identified a `Beach' as being adjacent to a `City'.
    \end{minipage}
    \caption{Live Earth observation~2: Argentina coast (47.80°S, 65.91°W).
    Uncorrected 10-bit$\rightarrow$8-bit capture processed autonomously onboard YAM-9. \textcopyright\ Loft Orbital 2026.}
    \label{fig:hero-argentina-raw}
    \label{fig:loft-raw-arg}
\end{figure*}

Two live images were captured and processed onboard (Figs.~\ref{fig:hero-toulouse} and~\ref{fig:hero-argentina-raw}), providing operational insight without ground segment intervention.

%% --- Hero: Toulouse raw ---

%% --- Hero: Argentina raw ---

These two detections represent, to our knowledge, the first instance of
a vision-language model performing autonomous multi-modal inference on
live-captured Earth imagery entirely onboard a satellite, with no
ground-in-the-loop connection.

% - - - - - - - - - - - - - - - - - - - - - - - - - - - (C.iv)
\subsubsection{Post-Processed}
\label{sec:results-new-orbit-retouched}
\label{sec:results-new-postproc}
\label{sec:results-new-retouched}

Early in-orbit data collection revealed that the newly deployed camera pipeline presents 10-bit RGB data to 8-bit processing routines which compressed the dynamic range and yielded visually dark images (Fig.~\ref{fig:loft-raw-arg}). Post-processing techniques were applied to recover dynamic range by extending truncated pixel values to the full 8-bit range, producing the post-processed images in Table~\ref{fig:hero-retouched}.

%% --- Hero: Retouched (Toulouse + Argentina combined) ---
\begin{table*}[!htb]
    \centering
    \begin{tabular}{@{}m{0.30\textwidth}@{\hspace{8pt}}p{0.66\textwidth}@{}}
        \toprule
        \textbf{Image} & \textbf{Detection (post-processed)} \\
        \midrule
        \includegraphics[width=0.30\textwidth]{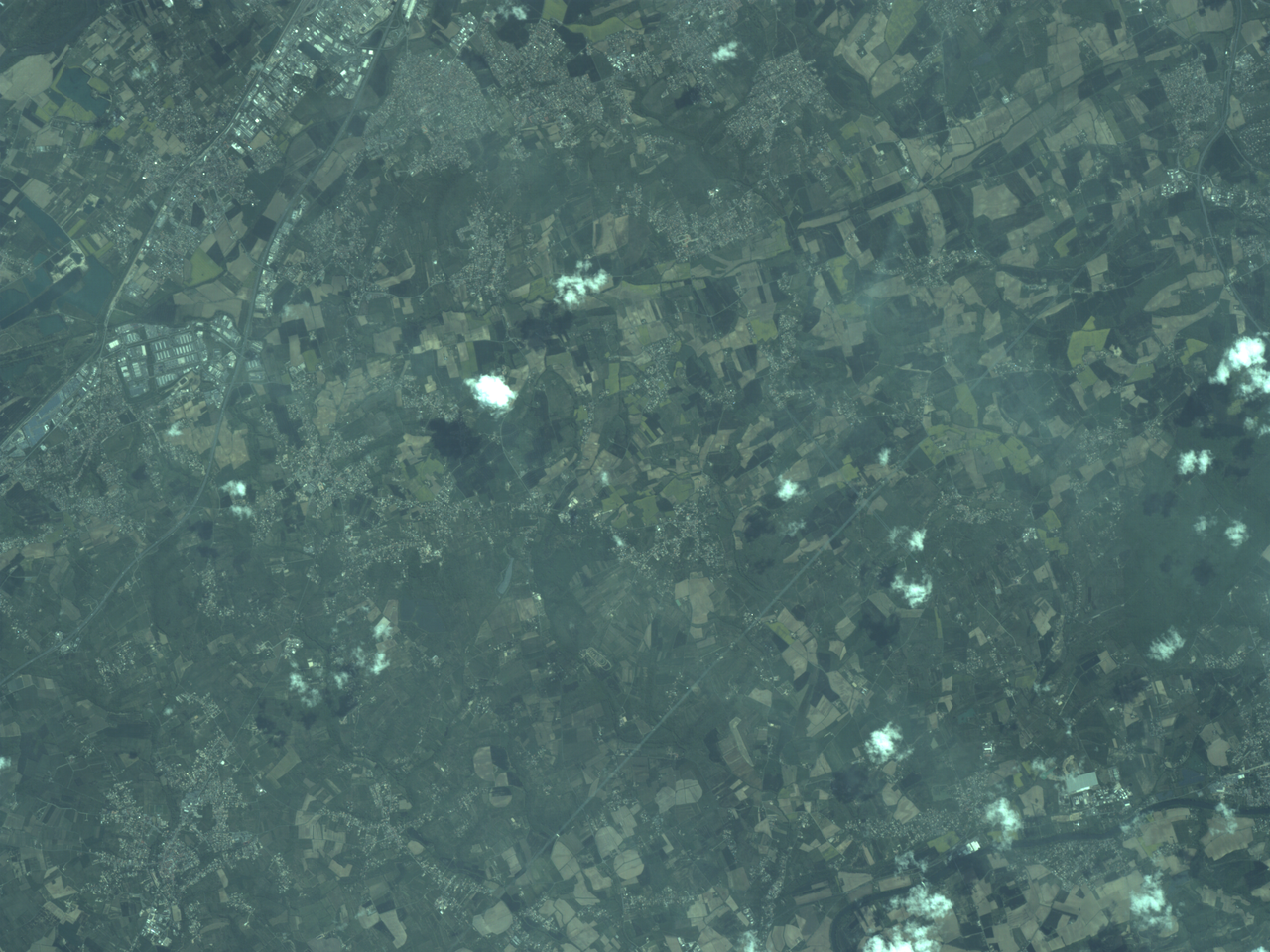} & \footnotesize
        \textit{Toulouse, France (43.76°N, 1.38°E). Label: Residential.}
        The image shows a densely populated area with a complex network of roads and buildings. There are numerous residential areas with distinct patterns of development, likely consisting of houses and apartment buildings. The terrain appears to be relatively flat, with some areas of green suggesting parks or undeveloped land interspersed among the built-up areas. There are also some scattered clouds in the sky. The overall impression is of a highly urbanized region with a significant residential population. \\[4pt]
        \includegraphics[width=0.30\textwidth]{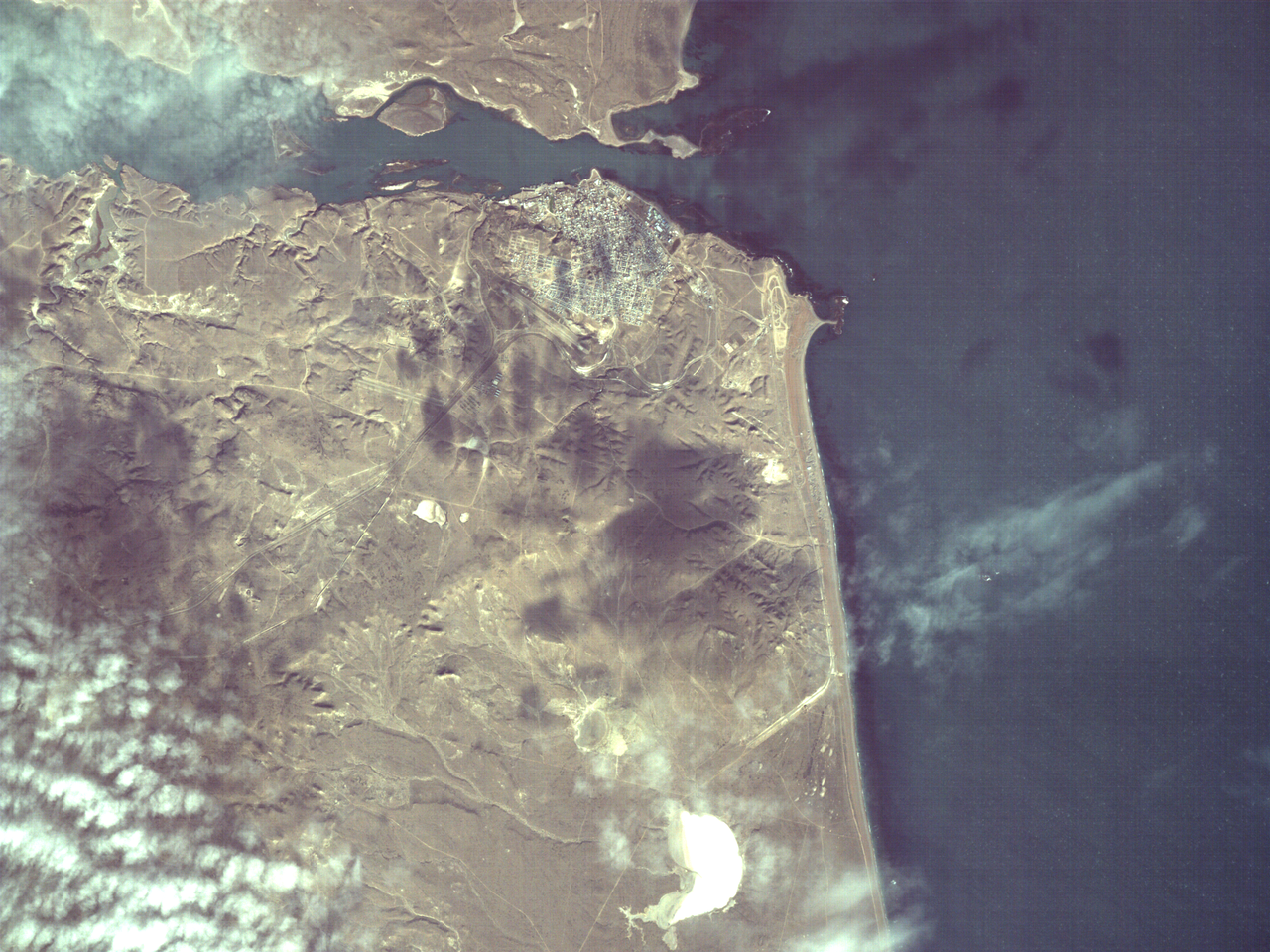} & \footnotesize
        \textit{Argentina coast (47.80°S, 65.91°W). Label: Beach.}
        The image shows a coastal area with a sandy beach extending along the coastline. The beach is bordered by a rocky shoreline and a mountainous region. There is a body of water, likely the ocean, visible in the background. A small, light-colored area, possibly a sandbar or a lagoon, is located near the center of the image. The presence of clouds in the sky suggests a somewhat overcast day. The overall scene appears to be a coastal landscape with a prominent beach feature. \\
        \bottomrule
    \end{tabular}
    \caption{Post-processed live captures re-run through NAVI-Orbital on the Flatsat after dynamic-range recovery. Compared with the raw onboard detections in Figs.~\ref{fig:hero-toulouse} and~\ref{fig:hero-argentina-raw}, the post-processed analyses recover finer geomorphic detail (rocky shoreline, mountainous terrain, clouds) while preserving the same core scene semantics. \textcopyright\ Loft Orbital 2026.}
    \label{fig:hero-retouched}
    \label{fig:hero-toulouse-retouched}
    \label{fig:hero-argentina-retouched}
    \label{fig:loft-retouched-tol}
    \label{fig:loft-retouched-arg}
\end{table*}

%%%%%%%%%%%%%%%%%%%%%%%%%%%%%%%%%%%%%%%%%%%%%%%%%%%%%%%%%%%%%%%%%%%%%%%%
%%  END staging Results section
%%%%%%%%%%%%%%%%%%%%%%%%%%%%%%%%%%%%%%%%%%%%%%%%%%%%%%%%%%%%%%%%%%%%%%%%

    \section{Discussion}

    \subsection{Hallucination and Reliability Analysis}
    \label{sec:hallucination}
    During ground benchmarking, 4 of 7{,}960 inferences (0.05\%) failed to produce a parseable label from the
    constrained set. Of these, 2 were resolved by the retry mechanism on a subsequent attempt, while 2 remained
    permanently unresolvable---in both cases the model invented labels outside the permitted set (``Golf Course'' and
    ``Park'', the latter emitted with spurious Markdown formatting). The retry mechanism (up to 3 attempts during
    ground evaluation, 10 during flight) successfully converges in $>$99.9\% of cases.

    The root cause of these failures is a tension between the model's generalist pre-training knowledge and the
    constrained label set: when an image exhibits high visual ambiguity or straddles defined class boundaries, the model may
    default to a label it considers more descriptively accurate rather than selecting the closest permitted option.
    This is a known limitation of constrained zero-shot classification with generalist VLMs and motivates the
    regex-based validation gate described in Section~\ref{NAVIArchitecture}.

    The dominant confusion patterns observed in Section~\ref{sec:results-AID} (Commercial District~$\rightarrow$~Residential, Water Reservoir~$\rightarrow$~Storage Tanks, Railway Station~$\rightarrow$~Industrial; Fig.~\ref{fig:confusion_matrix}) are semantically meaningful in a way that is qualitatively different from typical CNN failure modes. A human reviewer working at the same resolution would be drawn toward the same misclassifications: commercial districts share rooftop-and-road signatures with residential areas, water reservoirs sit adjacent to storage infrastructure, and railway stations are embedded in industrial zones. This is evidence that the model operates at a scene-level abstraction matched to the operational triage task, and that its errors are diagnosable through the text descriptions it produces.

    A second source of evidence for scene-level robustness comes from the post-flight reanalysis of live captures (Section~\ref{sec:results-new-orbit-live}). The Argentina coastal scene, processed in its raw 10-to-8-bit-truncated form and in a dynamic-range-recovered post-processed form, converged on the same core scene semantics: a coastal area with a sandy beach, a body of water, and a settlement on a peninsula. The post-processed version added finer detail (rocky shoreline, mountainous terrain, cloud cover) without changing the operational triage outcome. For downlink-prioritization, the onboard description on degraded input is already sufficient to drive selective downlink decisions; the post-processed reanalysis enriches but does not contradict the onboard inference.

    \subsection{Limitations}
    NAVI-Orbital's dialogue agent was exercised with 16 scripted questions per experiment, spanning label enumeration, comparative analysis, and semantic retrieval over the accumulated classification log (Table~\ref{tab:dialogue-showcase}). Adversarial prompts, including contradictions, false statements, and questions for which no supporting evidence exists in the context window, were outside the scope of this demonstration. The dialogue capability demonstrated in this work should therefore be read as a feasibility result rather than as a robustness characterization.

    The in-orbit demonstration comprises two live captures from the YAM-9 imager (Section~\ref{sec:results-new-orbit-live}), together with onboard post-processing of preloaded and cross-validation imagery (Section~\ref{sec:results-new-orbit-preload}). The resulting sample is small, comparable in scale to analogous in-orbit ML demonstrations such as ~\cite{MateoGarciaetal2023}; additional captures and re-tasking experiments are planned for subsequent operations.

    The primary limitation of NAVI-Orbital is structural rather than empirical. The system deploys a stochastic vision-language model into a domain (autonomous flight software) where the prevailing engineering convention is deterministic, formally verifiable behavior. No retry mechanism, regex output gate, or sampling-temperature setting eliminates the possibility of hallucination; these mitigations reduce its frequency but do not bound it. Integration of this class of software into broader onboard autonomous operations, for example, tasking another spacecraft subsystem on the basis of a VLM output, therefore requires architectural treatment beyond what NAVI-Orbital currently implements: supervisor agents and deterministic fallbacks, depending on the safety envelope of the dependent action. The demonstration in this paper establishes that vision-language models can be deployed on a spacecraft and produce operationally useful outputs; production integration into autonomy loops is a separate engineering problem that the community has not yet solved.

    \subsection{Implications for Cognitive Spacecraft}

    NAVI-Orbital operationalizes a deployment philosophy in which adaptation to new observation targets is a prompt change rather than a model change. The 18-class curation of the AID benchmark used in this work (Section~\ref{sec:datasetgoogle}) was a label-list edit; no fine-tuning, no architecture modification, no new training data. By the same mechanism, the system could in principle be re-tasked to other targets without modifying the underlying model, though this has not been validated beyond the AID and Sentinel-2 evaluations reported here, and any such generalization is bounded by the model's pre-training distribution. For tasks that fall within that distribution, this approach avoids the adapt-retrain-revalidate-re-uplink cycle typical of onboard classifier updates. The same zero-shot capability extended to imagery that had not been seen by the system before, including uncorrected YAM-9 captures processed with hardware-accelerated GPU inference and no per-instrument tuning.

    Underlying this approach is the model's capacity for multi-modal contextual reasoning: rather than mapping pixels to a single label, the VLM fuses visual input with language to produce a description that names the scene, its constituents, and their relationships (Section~\ref{sec:results-new-orbit-live}). Distilling this semantic knowledge onboard, rather than recovering it through post-hoc ground processing, is the prerequisite for a spacecraft to respond to its environment contextually, rather than reporting raw sensor data.

    NAVI-Orbital's combined detection and dialogue capability changes the operational interface between operators and the spacecraft. Tasking a satellite to recognize a new feature has historically required writing command sequences, re-validating onboard software, and uplinking new binaries; a workflow inherited from the specialist-detector paradigm described in Section~\ref{sec:intro}. Under the NAVI-Orbital paradigm, re-targeting amounts to editing and uploading a new prompt.  This shortens the re-tasking cycle and broadens the set of potential task authors beyond those with specialized command-sequence expertise.

    The bandwidth implications of this approach are quantifiable. Detection outputs across the AID benchmark suite occupy 700 to 1060\,bytes each in JSON form, comprising the predicted label, a free-text scene description, and inference metadata. The corresponding full-resolution YAM-9 captures are roughly 9 to 14\,MB for the live in-flight imagery (Section~\ref{sec:results-new-orbit-live}), placing the text summary roughly four orders of magnitude below its source frame in data volume. This four-orders-of-magnitude ratio amounts to a form of \emph{semantic compression}: rather than compressing pixels, the spacecraft compresses meaning, downlinking the textual content of an acquisition while discarding the visual data that is not needed to triage it. A spacecraft that downlinks a structured description of every acquisition, and accepts an operator follow-up to send full imagery only for high-value scenes, inverts the conventional acquire-then-downlink-everything bandwidth profile. This responds to the operator-capacity and downlink-bottleneck motivations identified in Section~\ref{sec:intro} directly, without requiring new ground-segment infrastructure or onboard storage growth.

    \subsection{Engineering Lessons Learned}
    \label{sec:lessons-learned}
    Four engineering lessons distilled from the NAVI-Orbital deployment campaign are worth recording for groups attempting similar work.

    \paragraph{Sub-byte quantization at the 4B-parameter scale was viable for this task}
    Gemma~3 at 4B parameters and 4-bit precision is materially smaller than current frontier vision-language models, yet performed adequately on this task: 88.16\% accuracy on the curated AID benchmark (Section~\ref{sec:results-AID}), and Flatsat-versus-orbit predictions matched on identical inputs (Section~\ref{sec:results-new-orbit-preload}).

    \paragraph{Bare-metal Python virtual environment was successfully deployed}
    Containerized deployment was the team's preferred operational model, but at integration time no container image tested provided reliable GPU access alongside the Python 3.12 and CUDA dependencies required by the inference pipeline. The flight configuration was therefore a bare-metal Python virtual environment on Comp2; an updated NVIDIA container release post-launch later validated containerized execution on the Flatsat (Section~\ref{sec:on-ground-testing}), but this was not uploaded. Container parity with bare-metal hardware acceleration on satellite-class edge processors should not be assumed at mission-integration time; planning a bare-metal fallback is worth the engineering cost.

    \paragraph{Pre-flight cross-validation on analogous-instrument imagery is informative}
    Cross-validation against imagery from instruments analogous to the flight payload provides evidence on whether the system generalizes to sensors representative of the flight instrument, before deployment on the flight model. For NAVI-Orbital this validation passed (Section~\ref{sec:results-new-cv}); a negative result would have been actionable, and our benchmark-set alone would not have surfaced it. The same zero-shot capability transferred beyond the analogous-instrument set to the two live YAM-9 captures themselves (Section~\ref{sec:results-new-orbit-live}), neither of which had been seen during development or ground testing.

    \paragraph{Persisting orchestration state to disk supports unsupervised recovery by design}
    Persisting orchestration state to disk between processing steps is the design pattern that allows NAVI-Orbital to recover autonomously. NAVI-Orbital implements this in LangGraph; the pattern transfers to any framework that serializes graph state between transitions. For autonomous in-orbit operations where ground intervention is not always available, this design choice represents defensive engineering.

    \section{Conclusions}
    To the best of the authors' knowledge, this work constitutes the first in-orbit demonstration of a vision-language model onboard a satellite, with autonomous multi-modal inference performed entirely onboard and no ground-in-the-loop connection during operations. Ground benchmarking on the 7{,}960-image curated AID benchmark established an 88.16\% accuracy baseline (Section~\ref{sec:results-AID}); in-orbit post-processing of preloaded inputs matched the ground predictions (Section~\ref{sec:results-new-orbit}); and two live YAM-9 captures, including uncorrected 10-bit imagery, were classified and described autonomously onboard with hardware-accelerated GPU inference, within a 20-second inter-capture cadence under flight thermal and power constraints, and without any fine-tuning for the YAM-9 imager (Section~\ref{sec:results-new-orbit-live}). Together, these results show that the natural-language-summary-and-dialogue pattern is a practical response to the downlink-bandwidth and operator-capacity bottlenecks identified in Section~\ref{sec:intro}, on currently available power-constrained edge processors.

    \subsection{Future Work}
    \label{sec:future-work}
    \begin{itemize}
        \item \textbf{Image segmentation and tracking.} Extending NAVI-Orbital beyond per-image classification, toward dense segmentation of features within an image and tracking of those features across successive captures, would enable applications such as change detection, plume monitoring, and dynamic event characterization. This capability has already been prototyped and tested, but was outside the scope of the present experiment; it will be evaluated in future iterations of NAVI.

        \item \textbf{Larger models and longer campaigns.} Running a more capable vision-language model on the same hardware platform, supported by a longer experimental campaign and a broader sample of acquired imagery, would clarify the accuracy and capability ceiling of the current envelope and surface failure modes that could not be exposed with the limited samples available.

        \item \textbf{Retrieval-augmented generation.} The next generation of NAVI incorporates retrieval-augmented generation (RAG), enabling the model to ground its outputs in retrieved context at inference time rather than relying solely on its parametric knowledge.
    \end{itemize}

    \section*{Acknowledgment}
    {The research was carried out at the Jet Propulsion Laboratory, California Institute of Technology, under a contract with the National Aeronautics and Space Administration, and supported by Loft Orbital's in-orbit edge compute platform. The authors extend their sincere gratitude to Paul Ramirez, Dr. Steve Chien and Dr. Tiago Vaquero from JPL for their mentorship and support. The authors also thank Pieter van Duijn of Loft Orbital, whose support made this in-orbit demonstration possible.}

    \bibliographystyle{IEEEtran}
    \bibliography{paper}

@inproceedings{gomez2024tackling,
  author    = {G{\'o}mez, Pablo and Meoni, Gabriele},
  title     = {Tackling the Satellite Downlink Bottleneck with Federated Onboard Learning of Image Compression},
  booktitle = {Proceedings of the IEEE/CVF Conference on Computer Vision and Pattern Recognition Workshops (CVPRW): AI4Space},
  pages     = {6809--6818},
  year      = {2024},
  doi       = {10.1109/CVPRW63382.2024.00674}
}

@article{giuffrida2021phisat,
  author    = {Giuffrida, Gianluca and Fanucci, Luca and Meoni, Gabriele and Bati{\v{c}}, Matej and Buckley, L{\'e}onie and Dunne, Aubrey and van Dijk, Chris and Esposito, Marco and Hefele, John and Vercruyssen, Nathan and Furano, Gianluca and Pastena, Massimiliano and Aschbacher, Josef},
  title     = {The {$\Phi$}-Sat-1 Mission: The First On-Board Deep Neural Network Demonstrator for Satellite Earth Observation},
  journal   = {IEEE Transactions on Geoscience and Remote Sensing},
  volume    = {60},
  pages     = {1--14},
  year      = {2022},
  publisher = {IEEE},
  doi       = {10.1109/TGRS.2021.3125567}
}

@techreport{google2025gemma,
  author    = {{Gemma Team}},
  title     = {Gemma 3 Technical Report},
  institution = {Google DeepMind},
  year      = {2025},
  note      = {arXiv:2503.19786}
}

@article{liu2024remoteclip,
  author    = {Liu, Chenyang and Zhang, Jiafan and Chen, Keyan and Wang, Man and Zou, Zhengxia and Shi, Zhenwei},
  title     = {RemoteCLIP: A Vision Language Foundation Model for Remote Sensing},
  journal   = {IEEE Transactions on Geoscience and Remote Sensing},
  year      = {2024},
  publisher = {IEEE}
}

@inproceedings{kuckreja2024geochat,
  author    = {Kuckreja, Kartik and Danish, Muhammad and Nasir, Muzammal and Das, Abhijit and Khan, Salman and Khan, Fahad Shahbaz},
  title     = {GeoChat: Grounded Large Vision-Language Model for Remote Sensing},
  booktitle = {Proceedings of the IEEE/CVF Conference on Computer Vision and Pattern Recognition (CVPR)},
  year      = {2024}
}

@article{xia2017aid,
  author  = {Xia, Gui-Song and Hu, Jingwen and Hu, Fan and Shi, Baoguang and Bai, Xiang and Zhong, Yanfei and Zhang, Liangpei and Lu, Xiaoqiang},
  title   = {AID: A Benchmark Data Set for Performance Evaluation of Aerial Scene Classification},
  journal = {IEEE Transactions on Geoscience and Remote Sensing},
  volume  = {55},
  number  = {7},
  pages   = {3965--3981},
  year    = {2017},
  publisher = {IEEE}
}

@software{gerganov2023llamacpp,
  author = {Gerganov, Georgi},
  title = {llama.cpp: Port of Facebook's LLaMA model in C/C++},
  year = {2023},
  publisher = {GitHub},
  journal = {GitHub repository},
  url = {https://github.com/ggerganov/llama.cpp}
}

@software{langchain2024langgraph,
  author = {{LangChain Inc.}},
  title = {LangGraph: Build resilient language agents as graphs},
  year = {2024},
  publisher = {GitHub},
  journal = {GitHub repository},
  url = {https://github.com/langchain-ai/langgraph}
}

@article{chien2005eo1,
  author    = {Chien, S. and Sherwood, R. and Tran, D. and Cichy, B. and Rabideau, G. and Castano, R. and Davies, A. and Mandl, D. and Frye, S. and Trout, B. and Shulman, S. and Boyer, D.},
  title     = {Using Autonomy Flight Software to Improve Science Return on {Earth Observing One}},
  journal   = {Journal of Aerospace Computing, Information, and Communication (JACIC)},
  pages     = {196--216},
  month     = apr,
  year      = {2005}
}

@article{giuffrida2020cloudscout,
  title={CloudScout: A Deep Neural Network for On-Board Cloud Detection on Hyperspectral Images},
  author={Giuffrida, Gianluca and Diana, Luca and de Gioia, Francesco and Benelli, Gionata and Meoni, Gabriele and Donati, Massimiliano and Fanucci, Luca},
  journal={Remote Sensing},
  volume={12},
  number={14},
  pages={2205},
  year={2020},
  publisher={MDPI},
  doi={10.3390/rs12142205}
}

@article{mateogarcia2021towards,
  author    = {Mateo-Garcia, Gonzalo and Veitch-Michaelis, Joshua and Smith, Lewis and Oprea, Silviu Vlad and Schumann, Guy and Gal, Yarin and Baydin, At{\i}l{\i}m G{\"u}ne{\c{s}} and Backes, Dietmar},
  title     = {Towards global flood mapping onboard low cost satellites with machine learning},
  journal   = {Scientific Reports},
  volume    = {11},
  number    = {1},
  pages     = {7249},
  year      = {2021},
  publisher = {Nature Publishing Group},
  doi       = {10.1038/s41598-021-86650-z}
}

@article{MateoGarciaetal2023,
  author  = {Mateo-Garcia, Gonzalo and Veitch-Michaelis, Josh and Purcell, Cormac and Longepe, Nicolas and Reid, Simon and Anlind, Alice and Bruhn, Fredrik and Parr, James and Mathieu, Pierre Philippe},
  title   = {In-orbit demonstration of a re-trainable machine learning payload for processing optical imagery},
  journal = {Scientific Reports},
  year    = {2023},
  volume  = {13},
  doi     = {10.1038/s41598-023-34436-w}
}

@article{LabrecheMladenov2023,
  author  = {Labrèche, G. and Mladenov, T.},
  title   = {Open-Source Software in Space Operations},
  journal = {Space Education \& Strategic Applications},
  year    = {2023},
  volume  = {4},
  doi     = {10.18278/sesa.4.1.2}
}

@INPROCEEDINGS{Wijata2024,
  author={Wijata, Agata M. and Lakota, Tomasz and Cwiek, Marcin and Ruszczak, Bogdan and Gumiela, Michal and Tulczyjew, Lukasz and Bartoszek, Andrzej and Longépé, Nicolas and Smykala, Krzysztof and Nalepa, Jakub},
  booktitle={IGARSS 2024 - 2024 IEEE International Geoscience and Remote Sensing Symposium}, 
  title={Intuition-1: Toward In-Orbit Bare Soil Detection Using Spectral Vegetation Indices}, 
  year={2024},
  volume={},
  number={},
  pages={1708-1712},
  doi={10.1109/IGARSS53475.2024.10640702}}

@article{justo2025hyperspectral,
  title={Hyperspectral Image Segmentation for Optimal Satellite Operations: In-Orbit Deployment of 1D-CNN},
  author={Justo, Jon Alvarez and Langer, Dennis D. and Berg, Simen and Nieke, Jens and Ionescu, Radu Tudor and Kjeldsberg, Per Gunnar and Johansen, Tor Arne},
  journal={Remote Sensing},
  volume={17},
  number={4},
  pages={642},
  year={2025},
  publisher={MDPI},
  doi={10.3390/rs17040642}
}

@article{zhang2022expandable,
  title={Expandable On-Board Real-Time Edge Computing Architecture for Luojia3 Intelligent Remote Sensing Satellite},
  author={Zhang, Zhiqi and Qu, Zhuo and Liu, Siyuan and Li, Dehua and Cao, Jinshan and Xie, Guangqi},
  journal={Remote Sensing},
  volume={14},
  number={15},
  pages={3596},
  year={2022},
  publisher={MDPI},
  doi={10.3390/rs14153596}
}

@inproceedings{dt-spaceops-2025, title={Flight of Dynamic Targeting on CogniSAT-6 - Update}, author={Chien, Steve and Zilberstein, Itai and Candela, Alberto and Rijlaarsdam, David and Perrocheau, Amaury and Dunne, Aubrey and Hendrix, Tom and Grauc, Oriol Cort{\'e}s and i Mestrec, Alexandre Gol and Bovec, Manel Pedra and Aragon, Oriol and Miquel, Juan Puig}, booktitle={Proceedings of the 18th International Conference on Space Operations}, year={2025}}

@misc{boozallen2024generative,
  author = {{Booz Allen Hamilton}},
  title = {Booz Allen Deploys the Power of Generative AI in Space},
  year = {2024},
  month = {August},
  note = {Press Release. Available at \url{https://newsroom.boozallen.com/news-releases/news-release-details/booz-allen-deploys-power-generative-ai-space/}},
}

@misc{meta2025spacellama,
  author = {{Meta} and {Booz Allen Hamilton}},
  title = {Space Llama: Meta's Open Source AI Model Is Heading Into Orbit},
  year = {2025},
  month = {April},
  note = {Meta Newsroom. Available at \url{https://about.fb.com/news/2025/04/space-llama-metas-open-source-ai-model-heading-into-orbit/}},
}

@misc{mousist2025astrea,
  title={ASTREA: Introducing Agentic Intelligence for Orbital Thermal Autonomy},
  author={Mousist, Alejandro D.},
  year={2025},
  eprint={2509.13380},
  archivePrefix={arXiv},
  primaryClass={cs.RO},
  url={https://arxiv.org/abs/2509.13380}
}

@article{kurt2026quantization,
  title={Which Quantization Should I Use? A Unified Evaluation of llama.cpp Quantization on Llama-3.1-8B-Instruct},
  author={Kurt, Uygar},
  journal={arXiv preprint arXiv:2601.14277},
  year={2026}
}

@inproceedings{dettmers2022optimizers,
  title={8-bit Optimizers via Block-wise Quantization},
  author={Dettmers, Tim and Lewis, Mike and Shleifer, Sam and Zettlemoyer, Luke},
  booktitle={9th International Conference on Learning Representations (ICLR)},
  year={2022}
}

@article{li2026rscovlm,
  title={Co-Training Vision Language Models for Remote Sensing Multi-task Learning},
  author={Li, Qingyun and Ma, Shuran and Luo, Junwei and Yu, Yi and Zhou, Yue and Wang, Fengxiang and Lu, Xudong and Wang, Xiaoxing and He, Xin and Chen, Yushi and Yang, Xue},
  journal={Remote Sensing},
  volume={18},
  number={2},
  pages={222},
  year={2026}
}

@inproceedings{jakubik2025terramind,
  title={TerraMind: Large-Scale Generative Multimodality for Earth Observation},
  author={Jakubik, Johannes and Yang, Felix and Blumenstiel, Benedikt and Scheurer, Erik and Sedona, Rocco and Maurogiovanni, Stefano and Bosmans, Jente and Dionelis, Nikolaos and Marsocci, Valerio and Kopp, Niklas and Ramachandran, Rahul and Fraccaro, Paolo and Brunschwiler, Thomas and Cavallaro, Gabriele and Bernabe-Moreno, Juan and Long{\'e}p{\'e}, Nicolas},
  booktitle={Proceedings of the IEEE/CVF International Conference on Computer Vision (ICCV)},
  pages={7383--7394},
  year={2025}
}

@article{wang2024agent,
  title={Agent AI with LangGraph: A Modular Framework for Enhancing Machine Translation Using Large Language Models},
  author={Wang, Jialin and Duan, Zhihua},
  journal={arXiv preprint arXiv:2412.03801},
  year={2024}
}

@article{sapkota2025langchain,
  title={LangChain vs. LangGraph vs. LangSmith: Taxonomies of Agentic AI Toolchains for End-to-End Orchestration},
  author={Sapkota, Ranjan and Shrestha, Rashik and Rijal, Madhav and Karkee, Manoj},
  journal={TechRxiv},
  year={2025}
}

@article{zhu2026orchestration,
  title={LLM-Based Multi-Agent Orchestration: A Survey of Frameworks, Communication Protocols, and Emerging Patterns},
  author={Zhu, Yiwen and Liu, Lihe and Yu, Jiaqian and Zhang, Di},
  journal={Preprints},
  year={2026}
}

@misc{loftorbital_imagery,
  author = {{Loft Orbital Inc.}},
  title  = {{Loft Orbital} Satellite Imagery},
  year   = {2026},
  note   = {Proprietary satellite imagery provided by Loft Orbital for this study. See \url{https://www.loftorbital.com}.}
}

@misc{zanaga2022esa,
  author    = {Zanaga, D. and Van De Kerchove, R. and Daems, D. and De Keersmaecker, W. and Brockmann, C. and Kirches, G. and Wevers, J. and Cartus, O. and Santoro, M. and Fritz, S. and Lesiv, M. and Herold, M. and Tsendbazar, N.E. and Xu, P. and Ramoino, F. and Arino, O.},
  title     = {{ESA WorldCover 10\,m 2021 v200}},
  year      = {2022},
  doi       = {10.5281/zenodo.7254221},
  publisher = {Zenodo}
}

@article{white2023prompt,
  title={A Prompt Pattern Catalog to Enhance Prompt Engineering with ChatGPT},
  author={White, Jules and Fu, Quchen and Hays, Sam and Sandborn, Michael and Olea, Carlos and Gilbert, Henry and Elnashar, Ashraf and Spencer-Smith, Jesse and Schmidt, Douglas C.},
  journal={arXiv preprint arXiv:2302.11382},
  year={2023}
}

@manual{nvidia2026tegrastats,
  author       = {NVIDIA-Corporation},
  title        = {Tegrastats Utility},
  organization = {NVIDIA},
  year         = {2026},
  note         = {NVIDIA Jetson Linux Developer Guide. Last updated Jan 16, 2026},
  url          = {https://docs.nvidia.com/jetson/archives/r36.4.4/DeveloperGuide/AT/JetsonLinuxDevelopmentTools/TegrastatsUtility.html}
}

@manual{nvidia2026orinpower,
  author       = {NVIDIA-Corporation},
  title        = {Jetson Orin Nano Series, Jetson Orin NX Series and Jetson AGX Orin Series},
  organization = {NVIDIA},
  year         = {2026},
  note         = {NVIDIA Jetson Linux Developer Guide. Last updated Jan 16, 2026},
  url          = {https://docs.nvidia.com/jetson/archives/r36.4.4/DeveloperGuide/SD/PlatformPowerAndPerformance/JetsonOrinNanoSeriesJetsonOrinNxSeriesAndJetsonAgxOrinSeries.html#jetson-agx-orin-series}
}

@article{Zareianetal2021,
  author  = {Zareian, Alireza and Rosa, Kevin Dela and Hu, Derek Hao and Chang, Shih-Fu},
  title   = {Open-Vocabulary Object Detection Using Captions},
  journal = {2021 IEEE/CVF Conference on Computer Vision and Pattern Recognition (CVPR)},
  year    = {2021},
  pages   = {14388--14397},
  doi     = {10.1109/cvpr46437.2021.01416}
}

@inproceedings{Radfordetal2021,
  author    = {Radford, Alec and Kim, Jong Wook and Hallacy, Chris and Ramesh, Aditya and Goh, Gabriel and Agarwal, Sandhini and Sastry, Girish and Askell, Amanda and Mishkin, Pamela and Clark, Jack and Krueger, Gretchen and Sutskever, Ilya},
  title     = {Learning Transferable Visual Models From Natural Language Supervision},
  booktitle = {Proceedings of the 38th International Conference on Machine Learning (ICML)},
  series    = {Proceedings of Machine Learning Research},
  volume    = {139},
  pages     = {8748--8763},
  year      = {2021},
  publisher = {PMLR}
}

% end of commented-out appendix
\end{document}